\PassOptionsToPackage{table}{xcolor}
\documentclass{article}

\usepackage{PRIMEarxiv}

\usepackage[utf8]{inputenc} 
\usepackage[T1]{fontenc}    
\usepackage{hyperref}       
\usepackage{url}            
\usepackage{booktabs}       
\usepackage{amsfonts}       
\usepackage{amsmath}
\usepackage{nicefrac}       
\usepackage{microtype}      
\usepackage{lipsum}
\usepackage{fancyhdr}       
\usepackage{float}
\usepackage{multirow}
\usepackage{tikz}
\usetikzlibrary{positioning, arrows.meta, shapes.geometric, fit, backgrounds, calc, decorations.pathreplacing}
\usepackage{xcolor}
\usepackage{placeins}

\usepackage{natbib}
\definecolor{slate}{RGB}{112,128,144}
\usepackage[most]{tcolorbox}
\newtcolorbox{querybox}[1]{enhanced, breakable,
  colback=slate!6, colframe=slate!80!black,
  colbacktitle=slate!80!black, coltitle=white,
  fonttitle=\bfseries\small, title={#1},
  boxrule=0.5pt, arc=2pt,
  left=6pt, right=6pt, top=4pt, bottom=4pt,
  fontupper=\small}
\newcommand{\querytablestyle}{%
  \setlength{\aboverulesep}{0pt}%
  \setlength{\belowrulesep}{0pt}%
  \renewcommand{\arraystretch}{1.3}%
  \rowcolors{2}{slate!6}{white}}
\usepackage{graphicx}       
\graphicspath{{media/}}     
\usepackage[title]{appendix}

\usepackage{subcaption}

\title{Agents and Tokens: Multi-Agent Autonomous Reasoning over a Hydrodynamic Token Store
}

\author{
  Jinpai Zhao \\
  Oden Institute \\
  University of Texas at Austin \\
  Austin, Texas, USA\\
  \texttt{max.zhao@utexas.edu} \\
  \And
  Albert Cerrone \\
  CEEES \\
  University of Notre Dame \\
  Notre Dame, Indiana, USA\\
  \texttt{acerrone@nd.edu} \\
  \And
  Joannes Westerink \\
  CEEES \\
  University of Notre Dame \\
  Notre Dame, Indiana, USA\\
  \texttt{jjw@nd.edu} \\
  \And
  Clint Dawson \\
  Oden Institute \\
  University of Texas at Austin \\
  Austin, Texas, USA\\
  \texttt{clint.dawson@austin.utexas.edu} \\
}

\begin{document}
\maketitle

\begin{abstract}

Large language models and the agents built on them are the most capable reasoning machines ever built, and science is beginning to put them to work. Atmospheric and oceanic sciences are starting to see agentic integration, but coastal hydrodynamics remains largely untouched by AI. Simply, agents cannot easily read its data products without saturating their context windows. A time frame from a coastal model can hold hundreds of megabytes, and if the archive's chunking does not suit the query, the agent will spend more time downloading and compacting than gaining insights. After all, the archives were designed for storage, not for the fast, on-the-fly access agents need. Speech and music once hit the same wall. A minute of audio is millions of numbers, yet today's audio models hold conversations and write music. They ``learned" to do this by ``listening" to tokens.  Neural audio codecs turn waveforms into short sequences of discrete symbols that a model reads like words and decodes back into sound. Coastal water is a similar signal: dense, finely sampled, and repeatable. We are convinced it has an analogous compact native representation that preserves the salient physics. Tokens are that representation: small integers, light enough for an agent to sweep six months of a coastline in seconds. Agents can then do more than simple lookup-related tasks; they can form and test their hypotheses as they explore the tokenized representation efficiently. Herein, we build this framework, using vector quantized variational autoencoders to compress tide level, the non-tidal residual, and surface wind speed into 16x16 grids of tokens along the US Atlantic and Gulf coasts. Agents traverse and explore these grids (e.g., tracking, spatiotemporal trend analysis) considerably faster than raw archives in both on-prem and off-prem modalities, while correctly answering several types of under-specified queries.

\end{abstract}


\section{Introduction}
Since AI became a household name in 2023 with the popularization of chat applications like OpenAI's ChatGPT and Google's Gemini, the general field of hydrodynamics has yet to be transformed by AI models in the same way fields like medical imaging and diagnostics and software engineering have been, where AI models are more commonplace and trusted.  This is not to suggest that the hydrodynamics community has been resistant to AI adoption.  In fact, judging by the number of machine-learning-centric hydrodynamics manuscripts published in the open literature, it seems that there would certainly be an appetite for imbuing automation in coastal modeling workflows.  It appears that the barrier precluding adoption is one of scale.  More specifically, the relevant scales of hydrodynamic model inputs and outputs (viz., spatial and temporal) and the sheer size of these inputs and outputs makes automating even simple hydrodynamic workflows a challenge.  This is not a question of computational expediency.  While hydrodynamic models like ADCIRC~\cite{Luettich1992ADCIRC} run more quickly and efficiently on GPUs, the limitations nevertheless persist.  This is because agentic activities that step beyond deterministic, pre-canned workflow execution require a sufficient number of degrees of freedom to prove / disprove their hypotheses.  This mandates that agents can interrogate relevant sources of data (be it atmospheric forcing or predicted total water levels) both rapidly and in a manner that does not overload their context windows.

Multi-agent systems (MAS) \cite{Yan2025BeyondSA} are a step in the right direction.  MAS incorporates multiple specialized agents to distribute the contextual burden, in a similar way that distributed computing relieves individual computing units.  Here, agents work together to satisfy an overarching user query, and for complex tasks, they generally do this more efficiently than single-agent systems (SAS) \cite{Guo2024LargeLM}.  Tool calling in SAS, for example, can saturate context windows relatively quickly \cite{Gao2025SingleagentOM, Johnson2025NaturalLT, Tran2025MultiAgentCM}, and this is most prominent with extensive tool calling lists.  The well known ``hallucination" occurs when an agent's context is saturated, and this can lead not only to inefficient tool calling, but also illogical / inaccurate output.  MAS effectively operates via a well-defined division of labor.  In this paradigm, specialized agents, each endowed with their own skills, tools, and directives (e.g., system prompts), distribute context throughout the system.  In relation to hydrodynamics, a specialized agent might be designed to source total water level observations from USGS, for example, or execute a hydrodynamics model altogether.  Of course, MAS has some drawbacks.  Coordinating the agents (a subject of the present work) requires thoughtful design while latency and token costs can make certain queries infeasible \cite{Xu2026RethinkingTV}.  Consequently, just as distributed-memory systems in high-performance computing (HPC) are not categorically superior to shared memory (this is entirely application-dependent), MAS should not be considered a wholesale improvement over SAS.

Coordination roughly follows one of three architectural patterns: hierarchical oversight, iterative debate, and sequential pipelines.  Hierarchical schemes leverage a supervisor of sorts, a central delegator that coordinates sub-tasks.  PANGAEA-GPT is an example of a hierarchical system in the geosciences as it exercises a centralized planner to route retrieval requests \cite{Pantiukhin2026AHM}.  Iterative debate architectures mimic what one might find in a boardroom, wherein experts (each assigned their own goals and responsibilities) exchange information for critique.  The desired result is some consensus.  It is noteworthy that this modality has been shown to reduce hallucinations \cite{Chan2024ChatEvalTB}.  Finally, sequential pipelines cascade context from agent(s) to agent(s).  AgentMD, an example of the sequential pipeline modality and inspiration for the present work, partitions its upstream curation of clinical calculators from its downstream execution of these calculators for diagnostic activities \cite{Jin2025AgentMD}.

In the present work, we pursue a hybrid of the hierarchical and sequential modalities.  Specifically, we lean on the layer execution graph (LEG) to scaffold the MAS.  A LEG is a constrained variant of the ubiquitous directed acyclic graph (DAG) in which agents within stacked execution layers fire sequentially.  The oversight comes from a ``graph'' specialist agent who designs the LEG from the user query.  Thereafter, the LEG is traversed from the top to bottom, ultimately ending with a ``reporter" agent that aligns the cascaded output to the query.  Graphs are not novel in MAS.  LangGraph \cite{langchain_langgraph} and GPTSwarm \cite{zhugegptswarm}, for example, both use computational graphs to coordinate agentic activities.  

The MAS LEG addresses context saturation, but it does not circumvent the aforementioned problem related to data access.  An agent embedded within a LEG charged with any tasks requiring search and/or exploration would be hard-pressed to fulfill its duties with conventional techniques.  Yes, cloud-optimized formats and virtual reference layers (e.g., Kerchunk and VirtualiZarr) have made retrieving data with time or location-based queries simpler and faster; however, they are largely ineffective for under-specified queries wherein multiple chunks of data (be it sliced in time or space) must be processed.  While this bottleneck impacts activities beyond AI, AI is particularly susceptible to it because agentic operations are frequently marked by on-the-fly revision.  Agents might reformulate their hypotheses during the course of a workflow, sometimes in unanticipated ways that could lead to unexpected download payloads.  This uncertainty means that under-specified queries might be more commonplace in the future, especially as these agentic technologies are more widely adopted.

Cherukuru \textit{et al.} \citep{Cherukuru2026SDE} provide a compelling solution by shifting the problem entirely into the latent space.  Specifically, they leverage embeddings proceeding from Meta's general-purpose vision transformer called DINOv3.  To obtain these embeddings, they inputted three-channel (RGB) images into DINOv3~\cite{Simeoni2025DINOv3} wherein the red channel was ERA5~\cite{Hersbach2020ERA5} mean sea-level pressure anomaly, the green channel was ERA5 10 m wind speed, and the blue channel was ERA5 total water column vapor.  They then populated a vector database with these embeddings (note that each embedding was a 768-length vector of 32-bit floats) and subsequently performed nearest-neighbor search to rapidly find analogs.  The utility here is that an embedding from the transformer encodes information about the ERA5 atmospheric quantities, albeit in a payload that is approximately 50x lighter than the input image.  Moreover, being a vector, the embedding can participate in rapid similarity-based analyses, facilitating search and exploration.

Like Cherukuru \textit{et al.} \citep{Cherukuru2026SDE}, we shift data access to the latent space, but we lean on tokens rather than embeddings.  Tokens, the codebook indices produced by vector quantization, are represented as discrete unsigned integers.  Herein, we consider ``grids" of tokens over geographically bounded extents so that they retain spatial coherence.  Like embeddings, tokens are extremely lightweight and can participate in rapid vector-based operations.  However, unlike continuous encoder embeddings (e.g., vision features), they can be decoded natively back to the physical space, meaning that reconstructions of the encoded physical quantity can be generated on-the-fly.  This does not mean that decoding is a necessary component of traversal, but that decoding in limited quantities can accentuate token patterns by grounding agentic latent space exploration with some physical relevance.  Additionally, these tokens can also be used to train a LLM from scratch or fine-tune an existing LLM with low-rank adaptation (LoRA) \citep{hu2022lora}.

This paper is organized into three sections.  First, a prototypical MAS is defined that includes specialized agents purpose-built to streamline typical hydrodynamic modeling workflows.  Second, a tokenization framework is proposed for efficiently traversing hydrodynamic model data products, with particular emphasis on total water levels and meteorological forcing.  This section details the tokenization method itself using vector quantized variational autoencoders (VQ-VAEs) \citep{vandenoord2017vqvae} and a token store with both temporal and spatial indexing that houses the resulting tokens.  The third section demonstrates the MAS and tokenization framework working together to facilitate agentic hydrodynamic modeling activities.

\section{A MAS Prototype}

\label{sec:prototype}

\subsection{Architecture}

The MAS prototype considered herein is composed of multiple ``specialist" agents.  A ``specialist" agent has its own skills and context window and a unique system prompt.  It cannot communicate directly with other specialists.  For all intents and purposes, each specialist would be isolated if it were not for the ``coordinator" agent, a general-purpose agent that constructs a LEG based on a user query.  Once the LEG is populated with agents, the graph is executed.  To relieve pressure on context windows, output from layers is summarized with a context compaction general-purpose agent.  This summarization is then sent to the agents in the proceeding layer (they do not see the direct output from agents in the preceding layer).  A final context consolidation is made after the last layer to ensure alignment with the user query.

The prototype is composed of the following specialist agents:

\begin{itemize}
    \item \texttt{NHC}: Supplies official tropical-cyclone guidance from the National Hurricane Center (NHC) including best tracks (HURDAT2~\cite{Landsea2013HURDAT2}) and current and archived advisories.  Some of the tools in this agent were adapted from OceanMCP (nhc-mcp)\cite{jisan2025oceanmcp}.
    \item \texttt{NOAA CO-OPS}: Supplies observed and predicted (viz., tide) water levels at stations situated along the US coast operated by the National Oceanic and Atmospheric Administration's (NOAA) Center for Operational Oceanographic Products and Services (CO-OPS).  Some of the tools in this agent were adapted from OceanMCP (coops-mcp) \cite{jisan2025oceanmcp}.
    \item \texttt{USGS}: Supplies survey-based high-water marks and Short-Term Network event data from the United States Geological Survey (USGS).
    \item \texttt{FEMA}: Supplies official flood insurance rate maps (FIRMs) and related digital products such as the National Flood Hazard Layer (NFHL) that show where inundation is regulated; mapped special flood hazard areas (SFHAs) identify locations subject to the base flood (often called the 1\% annual-chance flood), and base flood elevations (BFEs) give the reference flood height used on those maps for compliance and insurance rating.  This agent calls the Federal Emergency Management Agency's (FEMA) ArcGIS REST MapServer.
    \item \texttt{STOFS}: Supplies contour plots of NOAA's Global Storm Surge and Tide Operational Forecast System (STOFS-2D-Global)~\cite{Funakoshi2025STOFS} for a requested forecast cycle.  STOFS data are sourced from NOAA's S3 bucket on AWS housing operational outputs.
    \item \texttt{OSM}: Supplies Open Street Map (OSM) maps via OSM's tile server.  Both satellite and non-satellite maps can be sourced.  The primary output format of this agent is the .png file.
\end{itemize}

The prototype is also composed of five general-purpose agents:
\begin{itemize}
    \item \texttt{Graph Architect}: The ``coordinator" agent described above.  It receives the user query and constructs a LEG complete with specialist and other general-purpose agents.
    \item \texttt{Consolidator}: Consolidates the outputs of multiple specialist agents, summarizing them for forthcoming agents.  This agent always appears directly downstream of multi-agent layers.
    \item \texttt{Cross-Track Merge}: Consolidates the outputs of multiple upstream parallel tracks.  This agent always appears directly downstream of two or more parallel tracks.  Parallel tracks are independent of one another and designed by \texttt{Graph Architect}.
    \item \texttt{Image}: Analyzes the images output by specialist agents like \texttt{STOFS} and \texttt{OSM}.  This agent utilizes the image understanding capabilities of frontier LLMs.  Consequently, the primary function of this agent is to route images to the underlying LLM for interpretation.  This agent always appears directly downstream of specialist agents that produce images (viz., \texttt{STOFS} and \texttt{OSM}).
    \item \texttt{Reporter}: Generates a concise response addressing the particulars of the query.
\end{itemize}

At the heart of the MAS prototype is the \texttt{registry}, a manifest of sorts.  The fields below determine agent identity, how the \texttt{Graph Architect} selects and connects specialists in the LEG, and the allow-listed tool surface the runtime enforces:

\begin{itemize}
    \item \texttt{ID} - a stable key or alias
    \item \texttt{kind} - a string indicating whether the agent is a \texttt{specialist} (e.g., a tool caller) or \texttt{general-purpose}.
    \item \texttt{graph\_data\_class} - an additional designation for \texttt{specialist} agents indicating whether they are experts either in \texttt{observation}, \texttt{hypothetical}, or \texttt{forecast} - centric queries.
    \item \texttt{system\_prompt} - long-form instructions for agent (e.g., what it is, what APIs it may reason about, how it should behave, tool-calling logic)
    \item \texttt{router\_capabilities} - condensed version of \texttt{system\_prompt} that the \texttt{Graph Architect} uses to ascertain agent capabilities for proper \textit{routing}
    \item \texttt{tool\_names} and \texttt{openai\_tools} --- jointly define the allowlisted tool surface: \texttt{openai\_tools} supplies the schema the LLM sees, while \texttt{tool\_names} is what the server enforces at execution time before routing a call to the appropriate dispatch backend.
\end{itemize}

\begin{figure}
    \centering
    \includegraphics[width=0.8\linewidth]{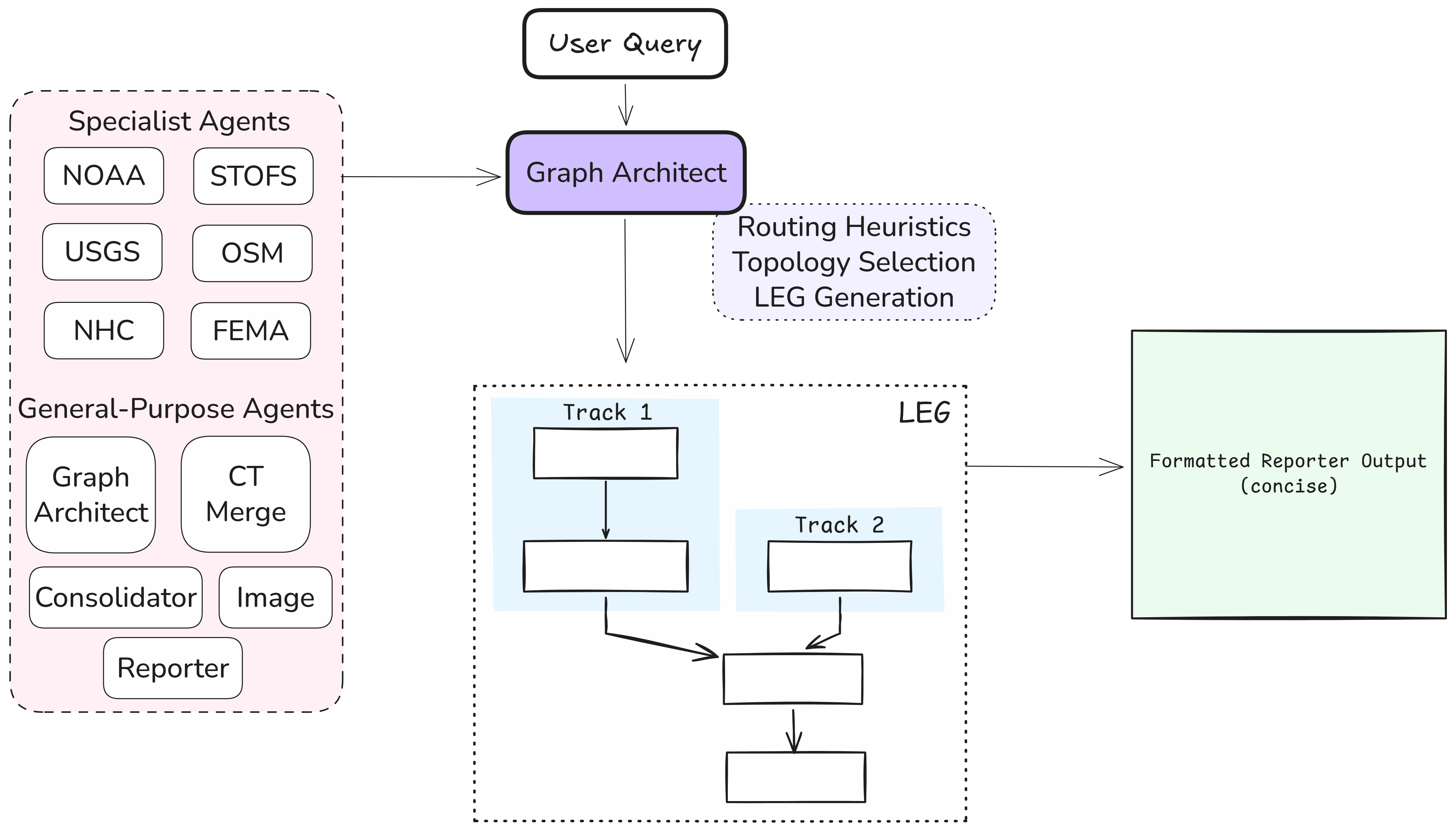}
    \caption{Diagram of MAS Prototype}
    \label{fig:mas_diagram}
\end{figure}

A diagram of the prototype's operation is given in Figure \ref{fig:mas_diagram}.

During execution, every specialist tool call is recorded including arguments, sourced data, and timing.  This is meant to support provenance-related queries.

\subsection{Demonstration}

To familiarize the reader with the general operation of the MAS prototype, we present four use-cases.  For an in-depth performance assessment of the prototype, the reader is directed to Appendix Section \ref{app:assessment}.

\subsubsection{Single Specialist Agent, Single Track}
\begin{figure}
    \centering
    \includegraphics[width=0.95\linewidth]{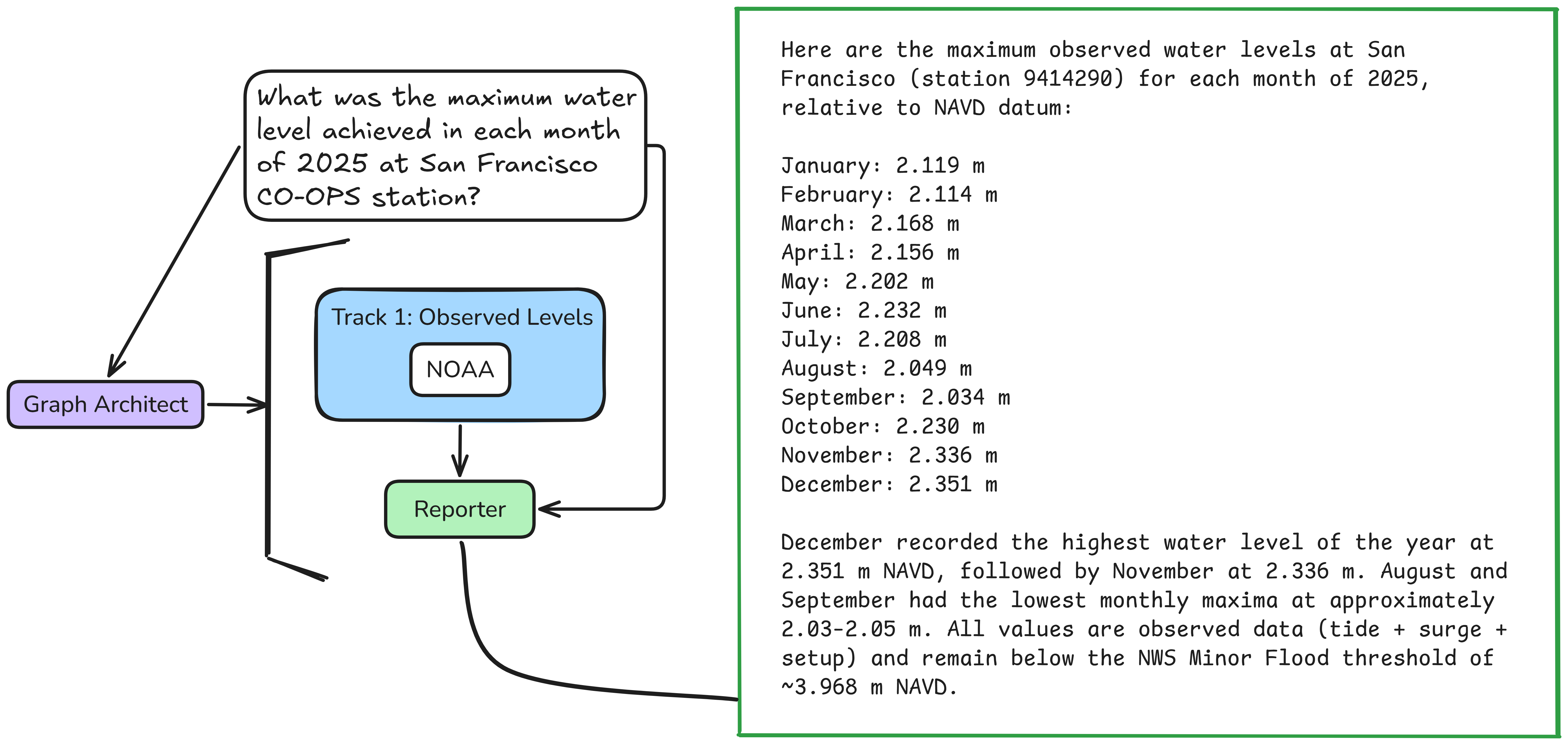}
    \caption{Example of one-track LEG with a single specialist agent.}
    \label{fig:single_specialist}
\end{figure}

We first consider a relatively simple user query: ``What was the maximum water level achieved in each month of 2025 at San Francisco CO-OPS station?''  To satisfy the user query, the \texttt{Graph Architect} generated a single-track LEG comprised of just one specialist agent and a reporter agent, Figure \ref{fig:single_specialist}.  During execution, the \texttt{NOAA CO-OPS} specialist agent called \texttt{noaa\_search\_stations} first to resolve ``San Francisco CO-OPS'' to station ID 9414290, then called \texttt{noaa\_get\_monthly\_water\_level\_stats} with the 2025 date range to fetch per-month aggregates. \texttt{noaa\_coops\_datagetter} was also called to source flood thresholds.  The agent presented the sourced data to the reporter, and the reporter subsequently responded to the query and provided some contextualization.

\subsubsection{Multiple Specialist Agents, Single Track}
\begin{figure}
    \centering
    \includegraphics[width=0.95\linewidth]{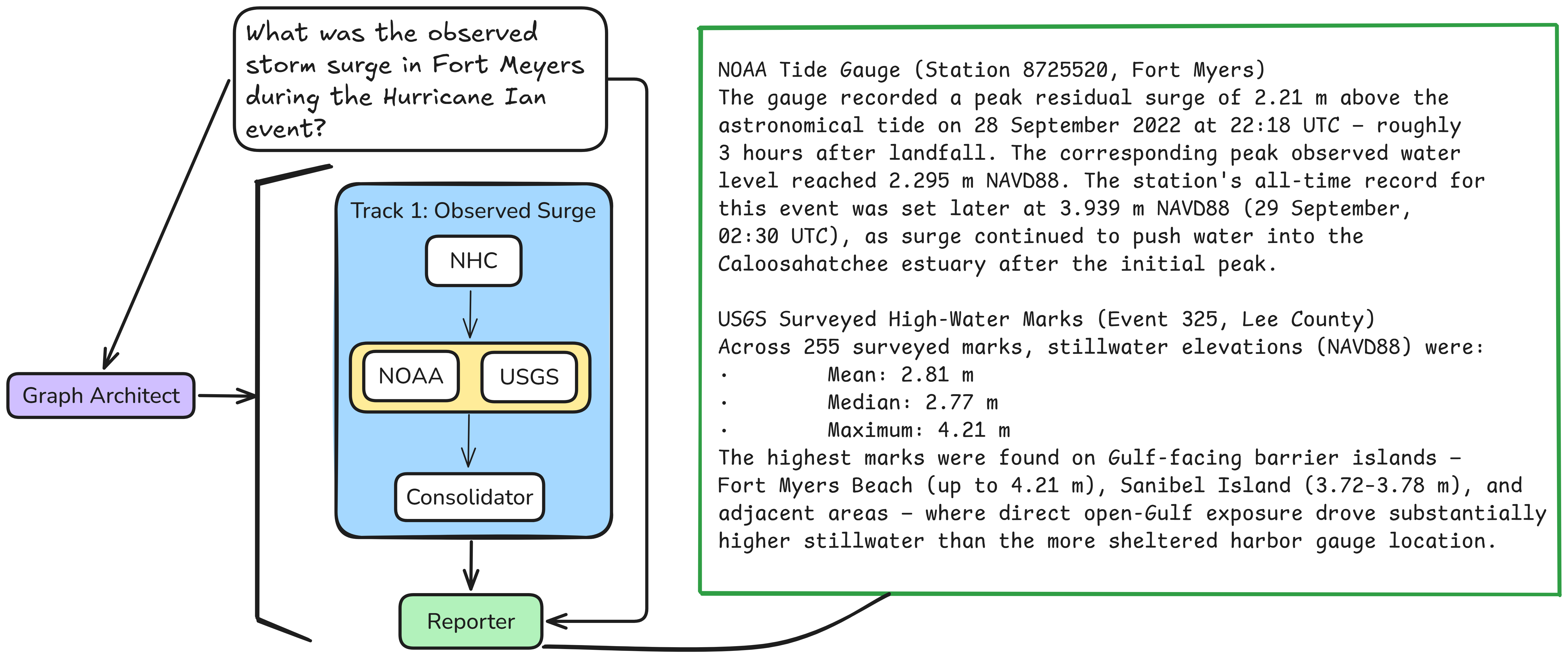}
    \caption{Example of one-track LEG with multiple specialist agents.}
    \label{fig:multiple_specialist}
\end{figure}

Next, we consider a slightly more complex query: ``What was the observed storm surge in Fort Myers during the Hurricane Ian event?''  To satisfy this query, the \texttt{Graph Architect} generated yet another single-track LEG, but scheduled three specialist agents within the layer, Figure \ref{fig:multiple_specialist}.  The \texttt{NHC} specialist agent was scheduled first because of the routing heuristics stored in the system prompt of the \texttt{Graph Architect}, specifically the prompt mentioning that the \texttt{NHC} agent should always precede other observation specialists when named storms are either directly invoked or otherwise needed to satisfy the query.  It called two tools, \texttt{nhc\_search\_storms} and \texttt{nhc\_get\_best\_track}, and passed the following output to the multi-agent layer below:

\begin{verbatim}

The storm made landfall near Fort Myers area around September 28, 2022 at
Category 4 intensity (max wind 140 kt, min pressure 937 hPa). The track 
shows the eye passed very close to Fort Myers:

2022-09-28 12:00:00 — lat 26.0, lon -82.7 — Category 5 (140 kt)
2022-09-28 18:00:00 — lat 26.6, lon -82.4 — Category 4 (135 kt)

The Fort Myers area (approximately 26.6°N, 81.9°W) was directly impacted
by the eyewall as Ian made landfall.

\end{verbatim}

The \texttt{NOAA CO-OPS} agent first called \texttt{noaa\_search\_stations} to source the CO-OPS station ID by the name ``Fort Myers''.  It then called \texttt{noaa\_find\_nearest\_stations} to find potentially other CO-OPS stations in the vicinity of Fort Myers (no other stations were found).  \texttt{noaa\_compute\_surge} was then called to compute the storm surge at Fort Myers in the interval Sept 28th 00 UTC to Oct 4th 00 UTC.  In parallel with the \texttt{NOAA CO-OPS} agent, the \texttt{USGS} agent first called \texttt{usgs\_stn\_resolve\_storm\_event} to source the event ID and then \texttt{usgs\_stn\_get\_hwms} to fetch the desired high water marks.  The \texttt{Consolidator} agent (which is a non-specialist) was invoked to consolidate the outputs of the \texttt{NOAA CO-OPS} and \texttt{USGS} agents.  Included in its output was a clarification about water level discrepancies that could have confused end-users:

\begin{verbatim}
Critical Discrepancy — Gauge vs. HWMs
The USGS HWMs (~4.2 m NAVD88) and CO-OPS gauge (2.30 m NAVD observed) report
substantially different peak elevations for the same storm at the same location.
This is expected and methodologically consistent:

CO-OPS 8725520 sits ~15 km upstream on the Caloosahatchee River. The estuary,
mangroves, and channel narrowing attenuate surge before it reaches the gauge. 
The gauge is an instrumented inland location, not an open-coast site. USGS HWMs
represent stillwater elevations at the open Gulf-facing coast (Fort Myers Beach, 
Sanibel, Pine Island) where surge piled directly against the barrier islands.
The USGS stillwater marks (4.2 m) represent the coastal open-water peak; the 
CO-OPS gauge (2.3 m) represents the in-river arrival of that surge after
propagation and attenuation. These datasets are complementary, not contradictory.
\end{verbatim}

This is an example of the MAS prototype's interpretation capabilities.  Not only did the \texttt{Consolidator} agent perceive a discrepancy, but it also reconciled it.  This is reflected in the last sentence of the output of the \texttt{Reporter} agent in Figure \ref{fig:multiple_specialist}.

\subsubsection{Parallel Tracks}
\begin{figure}
    \centering
    \includegraphics[width=0.9\linewidth]{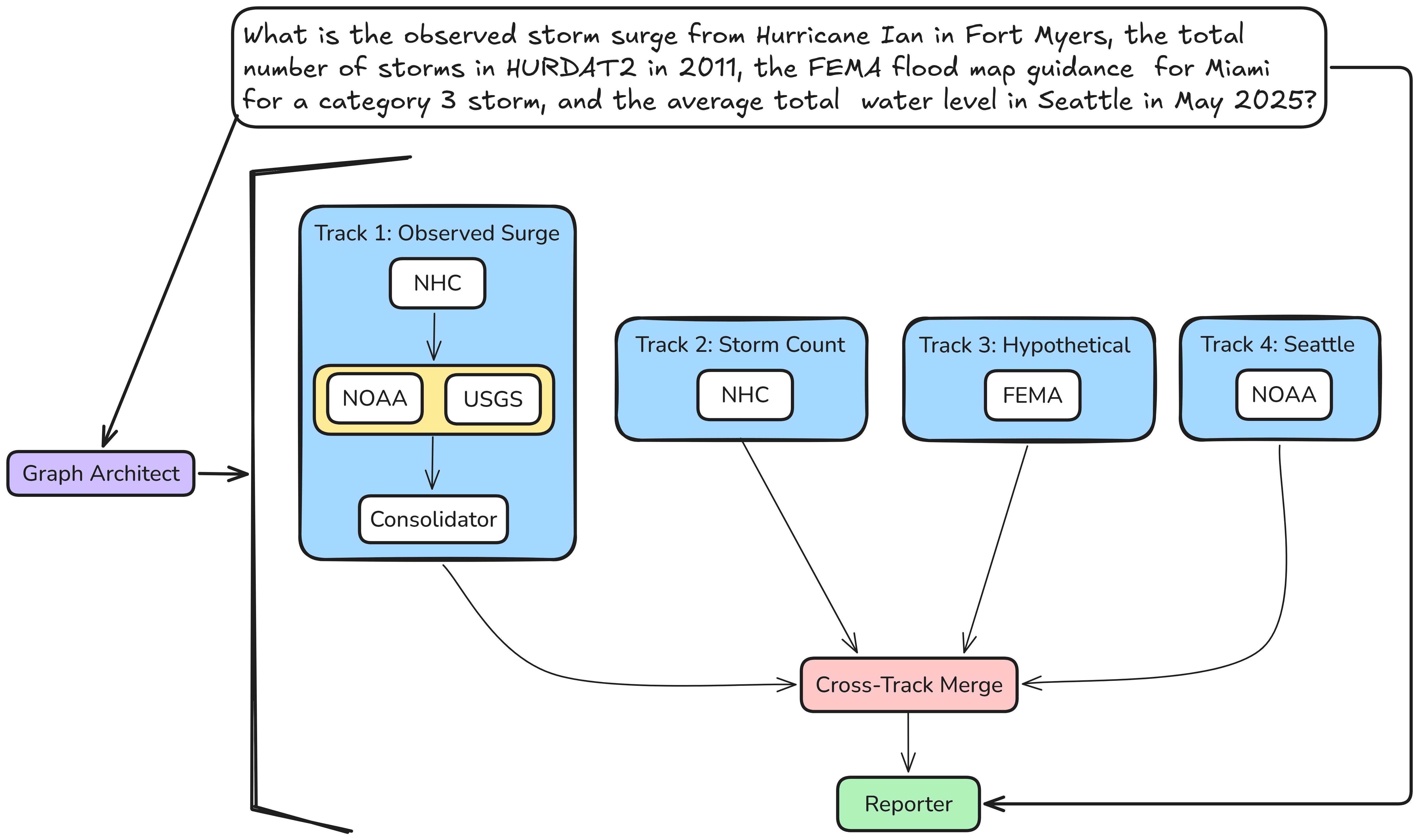}
    \caption{Example of four-track LEG.}
    \label{fig:four_tracks}
\end{figure}

We now consider the following multi-part query: ``What is the observed storm surge from Hurricane Ian in Fort Myers, the total number of storms in HURDAT2 in 2011, the FEMA flood map guidance for Miami for a category 3 storm, and the average total water level in Seattle in May 2025?"  The \texttt{Graph Architect} immediately allocated four parallel tracks, one for each of the four independent questions in the query, Figure \ref{fig:four_tracks}.  Note that the \texttt{NHC} agent appears in both Tracks 1 and 2, but the system properly spawns an agent for each.  In this case, the second query could have easily been addressed by the agent in Track 1; however, for more convoluted tasks, daisy-chaining queries (as opposed to segregating them in parallel tracks) can overload context windows and lead to degraded performance (e.g., neglecting a query altogether).  As shown in the output of the \texttt{Reporter} below, each query is addressed in a fairly concise manner:

\begin{verbatim}
**1. Observed storm surge from Hurricane Ian at Fort Myers**
Peak storm surge of 2.209 m above predicted tide was recorded at NOAA CO-OPS station
8725520 (Fort Myers), arriving at 22:18 UTC on 28 September 2022 — roughly three hours
after Ian made landfall at Cayo Costa as a Category 4 hurricane. This is the stillwater
surge component measured at the Fort Myers yacht basin. USGS high-water marks across Lee
County show higher peak totals of up to 4.20 m NAVD88 at coastal locations because those
marks include wave run-up and overwash in addition to the stillwater surge; the mean HWM
across 255 marks was 2.81 m.
**2. Total number of storms in HURDAT2 in 2011**
The Atlantic basin HURDAT2 database lists 20 named storms in 2011. Of those, 4 were major
hurricanes (Category 3 or higher): Irene (Cat 3), Katia (Cat 4), Ophelia (Cat 4), and 
Rina (Cat 3).
**3. FEMA flood map guidance for Miami for a Category 3 storm**
At the Miami point queried (25.7617° N, 80.1918° W), the effective FEMA NFIRM identifies
VE zones (coastal high-hazard areas with wave-action risk) and AE zones (inland/surge-
related Special Flood Hazard Areas). Mapped Base Flood Elevations span 2.74–4.57 m NGVD29
across these zones. These are the regulatory 1%-annual-chance ("100-year flood")
elevations, not storm-specific predictions. During an actual Category 3 landfall,
observed surge could substantially exceed these mapped BFEs depending on track, tide, and
forward speed.
**4. Average total water level in Seattle in May 2025**
The mean total water level at NOAA station 9447130 (Seattle) for May 2025 was 1.312 m
NAVD, based on 7 440 six-minute observations. Values ranged from -1.919 m to 3.076 m over
the full month.
\end{verbatim}

\subsubsection{Image Understanding}
\begin{figure}
    \centering
    \includegraphics[width=0.95\linewidth]{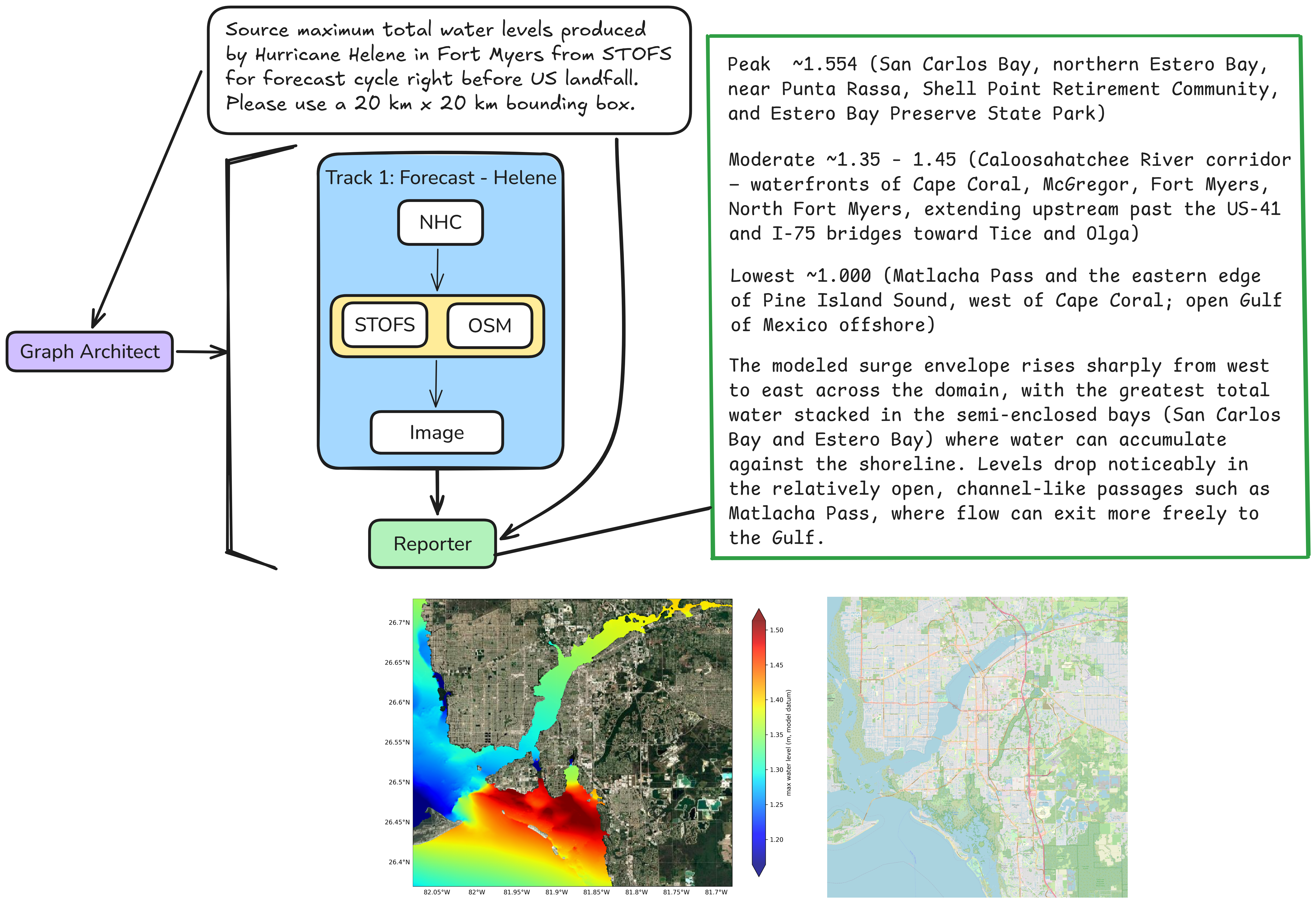}
    \caption{Example of image understanding (\copyright~OpenStreetMap contributors).}
    \label{fig:image_understanding}
\end{figure}

Finally, we consider the query: ``Source maximum total water levels produced by Hurricane Helene in Fort Myers from STOFS for forecast cycle right before US landfall.  Please use a 20 km x 20 km bounding box.''  This query invokes the \texttt{STOFS} agent directly, so the \texttt{Graph Architect} agent automatically stages the \texttt{OSM} agent in the same layer as \texttt{STOFS}, Figure \ref{fig:image_understanding}.  Additionally, because both agents in the layer generate images, the \texttt{Image} agent is placed directly downstream of them.

The \texttt{Image} agent is a form of consolidation, a potentially necessary one for outputs proceeding from high-resolution models like STOFS-2D-Global stored in monolithic archives.  Note that a frontier LLM could not possibly be expected to ingest millions of floating point numbers proceeding from an unstructured grid product like STOFS and establish spatial coherence.  Image understanding is a means to draw context out of STOFS output (via contour plots of total water level), effectively condensing hundreds of megabytes of data per time frame into a narrative complete with geographic references.  The LLM's base knowledge obviously contributes to this referencing, but the \texttt{Image} agent actually references a labeled OSM basemap together with the STOFS plot, meaning that reporter output can be highly localized.  This is evident in the \texttt{Reporter} agent's summary of maximum total water levels in the Fort Myers region, Figure \ref{fig:image_understanding}.

This particular demonstration ultimately motivates a token-based traversal scheme.  While the user query was well specified as it referenced a landfall location (Fort Myers) and time (cycle before Hurricane Helene's landfall), a more open-ended query would have stretched the limits of expedient agentic processing.  For example, if the query named multiple locations, several storms, and/or different hurricane seasons, the resulting image understanding calls would have quickly overloaded API expenditure budgets.

\section{Tokenization Framework}

The tokenization framework is composed of two inter-connected elements: (1) the tokenization effort itself that leverages VQ-VAEs to generate tokens of hydrodynamic model inputs and outputs and (2) a token store with an accompanying API to facilitate rapid search and exploration.

\subsection{Tokenization}

\subsubsection{Training VQ-VAEs}

A VQ-VAE is an autoencoder.  It compresses inputs via nearest-neighbor quantization in the latent space into a grid of discrete indices or tokens.  Note that these tokens, which are represented as unsigned integers, map to learned codebook vectors where $K$ is codebook size (i.e., number of unique token values) and $D$ is the latent dimension (i.e., codebook vector length).

Three VQ-VAEs were trained for this study: tide level, non-tidal residual (NTR), and surface wind.  Water levels were computed for the US Atlantic Seaboard and Gulf Coast with a JAX-based~\cite{jax2018github} shallow-water equation (SWE) solver developed specifically for this study.  It integrates continuity and momentum on a structured Arakawa C-grid~\cite{Arakawa1977UCLA} with explicit time stepping.  The tidal potential is calculated directly from the relative positions of the sun and moon using the same method embedded in ADCIRC~\cite{Luettich1992ADCIRC}.  The contribution of self-attraction and loading (SAL) is also added to the equilibrium tide height for the momentum equation.  Atmospheric forcing is provided by the Global Forecast System (GFS).  The model's topo-bathymetry and SAL are lifted from the ADCIRC-driven STOFS-2D-Global model~\cite{Funakoshi2025STOFS} operated by NOAA 4x daily.  Tidal boundary forcing is provided by TPXO~\cite{Egbert2002TPXO}.  Wetting and drying, finally, follows face-based wetting and drying on the C-grid (inspired by SFINCS~\cite{Leijnse2021SFINCS}).

\begin{figure*}[t]
  \centering
  \begin{subfigure}[t]{0.48\textwidth}
    \centering
    \includegraphics[width=\linewidth]{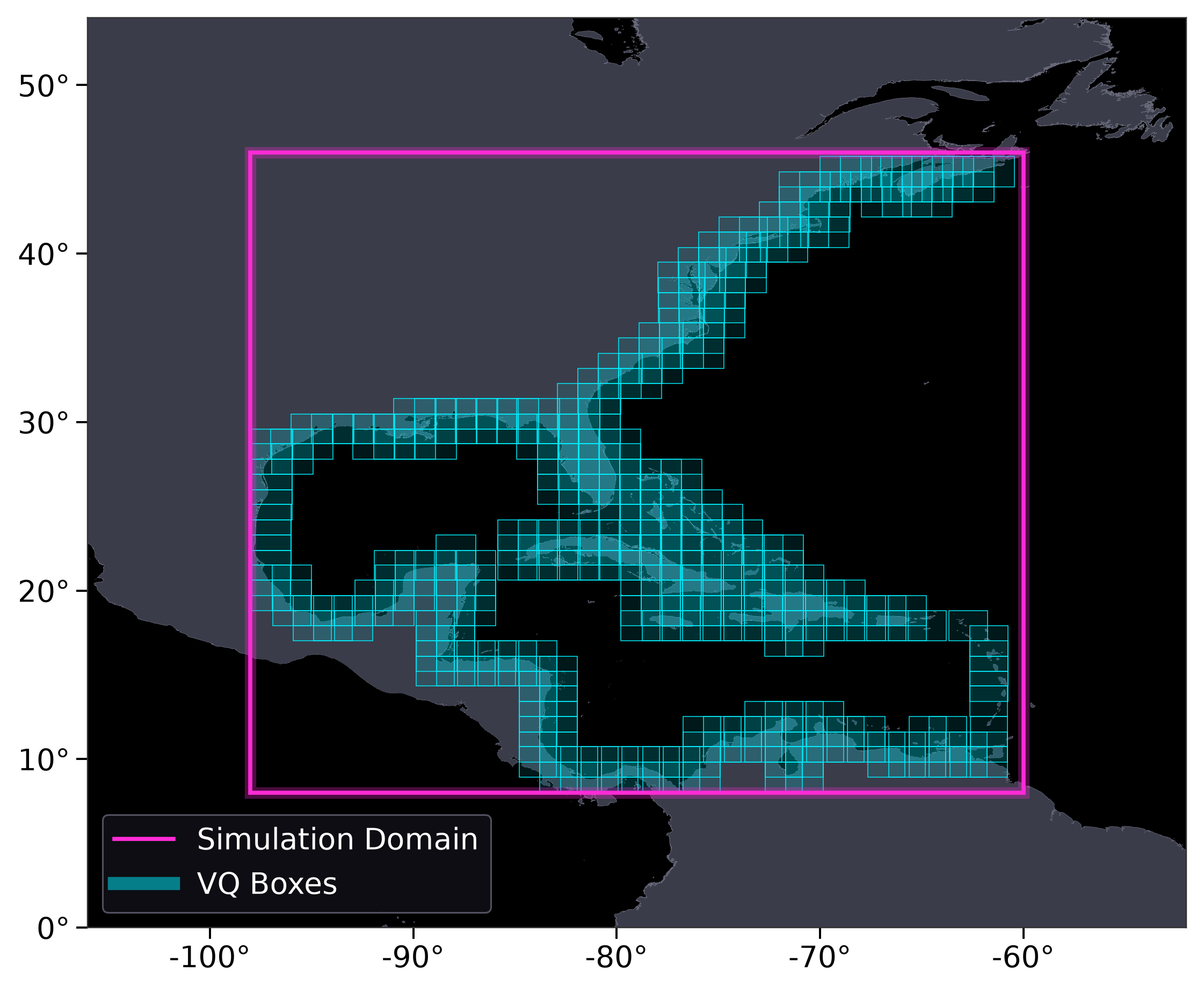}
    \caption{Simulation Domain and VQ Boxes}
    \label{fig:domain-vq}
  \end{subfigure}
  \hfill
  \begin{subfigure}[t]{0.48\textwidth}
    \centering
    \includegraphics[width=\linewidth]{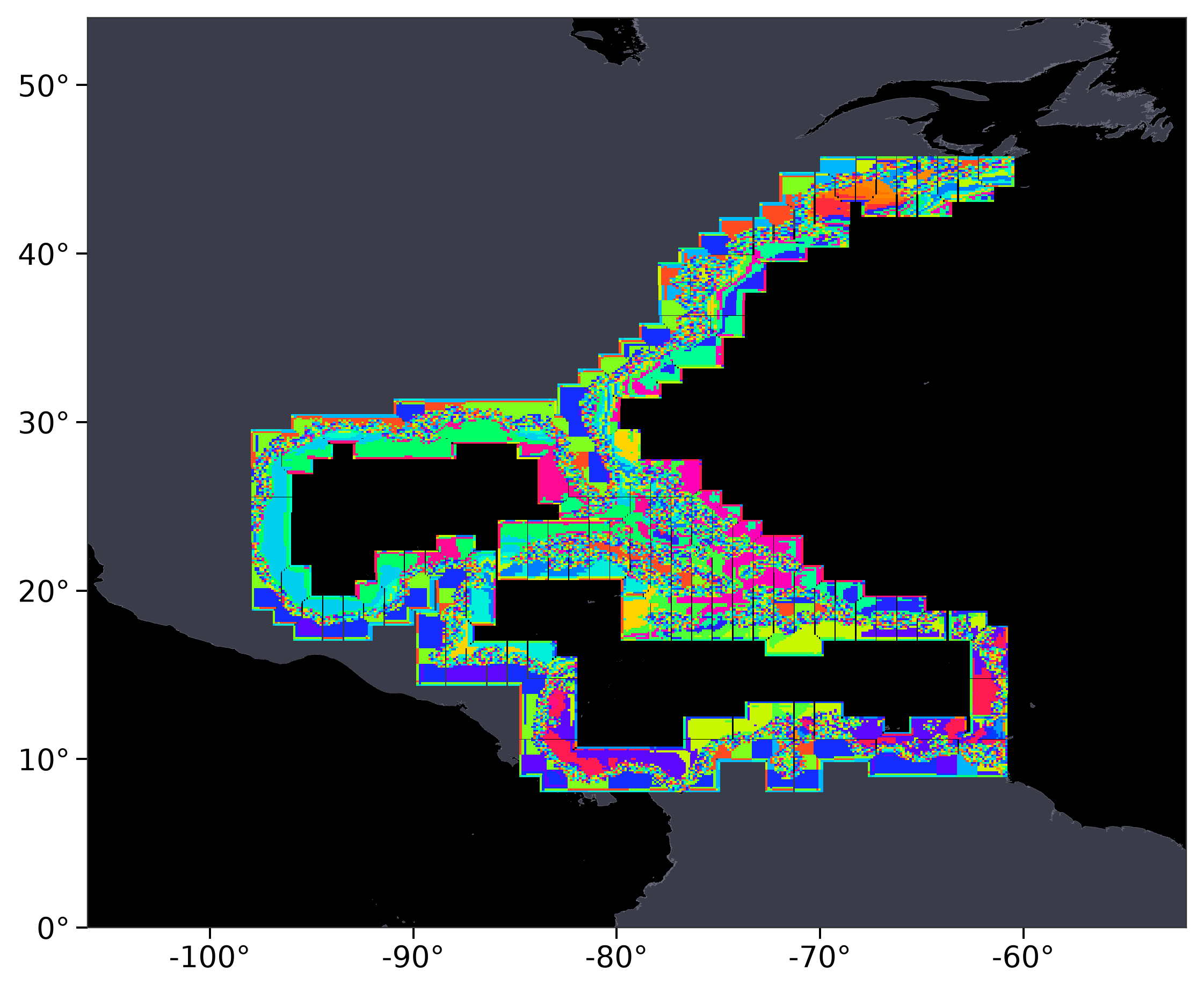}
    \caption{Tide Level Tokens}
    \label{fig:tide-tokens}
  \end{subfigure}
  \caption{Arrangement of VQ boxes in simulation domain (a) and tide level tokens from January 1, 2026 midnight UTC (b, each discrete color represents a token index from 0 to 255 inclusive).}
  \label{fig:token-layout}
\end{figure*}

The solver was run twice for the period January 1, 2026 - July 14, 2026 with 0.03125$^{\circ}$ grid resolution.  Predictions from the second half of June were reserved for validation while the July predictions were reserved for testing.  First, it was run without atmospheric forcing to produce predictions of tide level.  Thereafter, it was run with atmospheric forcing to produce predictions of total water level.  The NTR was found by subtracting the tide level from the total water level.  Coastlines were then discretized into 350 200-km boxes stepped every 100 km.  In the latent space, each box was represented as a $16 \times 16$ discrete grid of VQ (vector quantized) codes.  The simulation domain, these VQ boxes, and a sampling of tokens are shown in Figure \ref{fig:token-layout}.

The three VQ-VAEs encode $256 \times 256$ rasterized representations of tide level, NTR, and surface winds into $16 \times 16$ latent grids.  Each has its own codebook of $K$=256 vectors of dimension $D$=64.  The Tide and NTR autoencoders share the same CNN-VQ backbone with two-channel input (normalized tide level / NTR and wet/dry indicator).  The surface wind autoencoder also shares the same backbone, but with normalized 10 m eastward and northward velocity components in place of tide level / NTR and the wet/dry indicator.  Note that the surface wind VQ-VAE does not represent wind speed on land.  Each was trained over 40 epochs wherein, for each epoch, 4,000 samples were drawn with replacement from a pool of approximately 70,000 samples.  A constant learning rate of $2\times10^{-4}$ was applied.  For validation samples, the lowest mean absolute error (MAE) achieved was 1.48 cm, 0.67 cm, and 0.28 m/s for the tide level, NTR, and surface wind VQ-VAEs, respectively.  The trained VQ-VAEs were then applied to samples completely independent of the training / validation set from July 1, 2026 - July 14, 2026.  The decoded reconstructions were compared to  the original $256 \times 256$ rasters, and the differences were aggregated to calculate the error metrics shown in Table \ref{tab:vq-hybrid-jul}.

\begin{table}[t]
\centering
\caption{VQ-VAE reconstruction errors for period 1--14 July 2026 (350 boxes $\times$ 336 hours).  Errors measured as absolute and subsequently aggregated to arrive at the indicated metric.}
\label{tab:vq-hybrid-jul}
\begin{tabular}{lrrrrrrrrr}
\toprule
Track & Mean & RMSE & Bias & $Q_{25}$ & $Q_{50}$ & $Q_{75}$ & $Q_{95}$ & $Q_{99}$ & Max \\
\midrule
Tide (cm) & 1.36 & 3.28 & $-0.02$ & 0.25 & 0.75 & 1.25 & 4.75 & 12.75 & 268.1 \\
NTR (cm) & 0.54 & 1.23 & 0.10 & 0.25 & 0.25 & 0.75 & 1.75 & 4.75 & 514.9 \\
Wind (m\,s$^{-1}$) & 0.243 & 0.377 & $-0.013$ & 0.075 & 0.175 & 0.325 & 0.725 & 1.375 & 15.39 \\
\bottomrule
\end{tabular}
\end{table}

\begin{figure*}[t]
  \centering
  \includegraphics[width=0.9\linewidth]{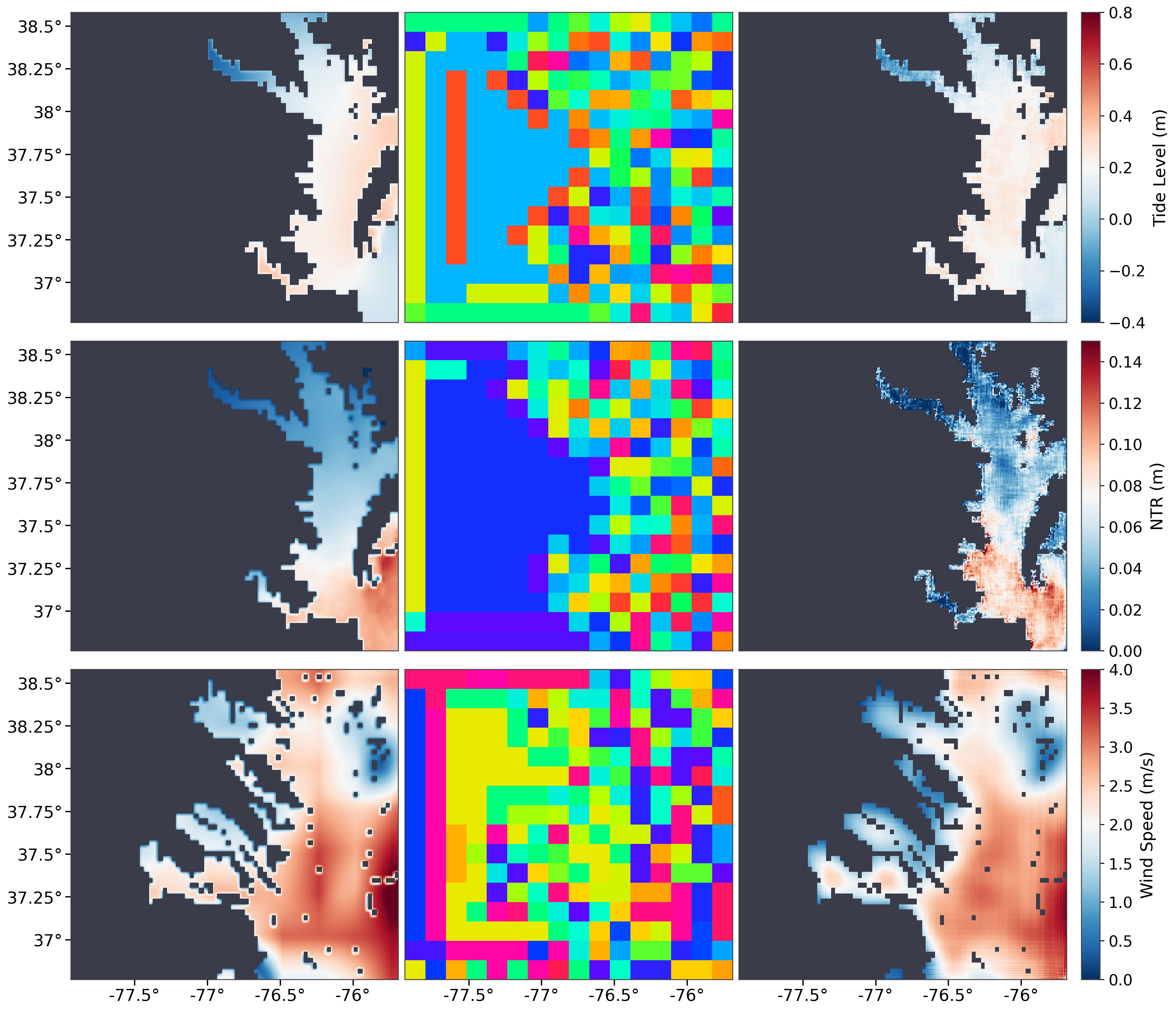}
  \caption{VQ-VAE encode/decode for one southern Chesapeake Bay box at July 7, 2026 21:00 UTC.
  Rows are tide level (top), NTR (middle), and wind speed (bottom).  Columns are the raw $256 \times 256$ rasters (left), the $16\times16$ codebook token grids (middle), and the decoded $256 \times 256$ reconstructions (right).}
  \label{fig:encode-decode}
\end{figure*}

Selected side-by-side comparisons of the ``raw" rasterized inputs and their decoded counterparts are given in Figure \ref{fig:encode-decode}.

\subsubsection{The Token Store and its Traversal}

The token grids are stored in a database dubbed the token store.  The token store is a compressed archive of discrete VQ codes with several entry points for under-specified queries.  Each grid or \texttt{frame} is tagged with a box index (an unsigned integer ranging from 0-349), a time index, and a track indicating the associated codebook (viz., \texttt{tide}, \texttt{ntr}, and \texttt{wind}).  The store has several lookup tables to facilitate resolving box and time indices with coordinates and datetimes, respectively.  For example, the location-based lookup maps each box index to a lon--lat bounding box, lon--lat box center, overlapping box indices, and indices into the source netCDF archive produced by the JAX-based code.  The time-based lookup, in turn, maps each time index to a UTC Unix timestamp so ISO datetimes can be snapped to the nearest hour in the store.  Indexes are also inverted to facilitate bulk queries.  For example, a spatial query maps a lon--lat bounding box to box indices, which can then be mapped to frame IDs for the time interval of interest.

\begin{figure*}[t]
  \centering
  \includegraphics[width=0.6\textwidth]{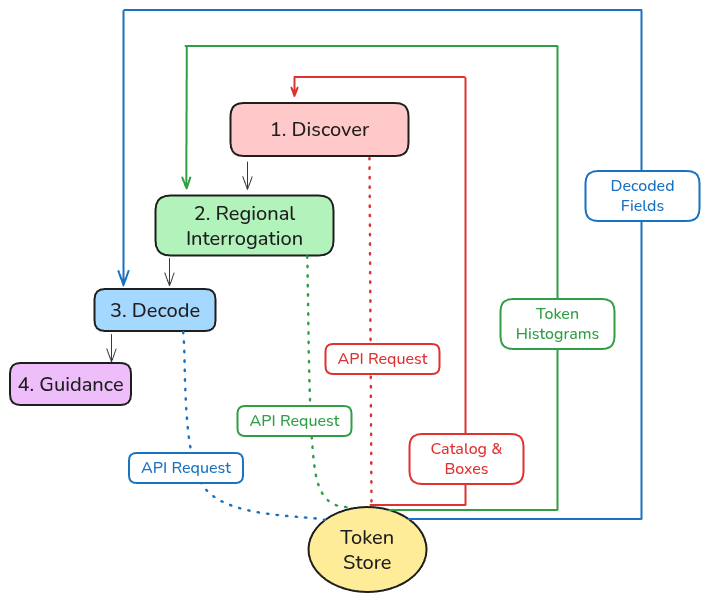}
  \caption{Prototypical activity of the \texttt{Token Store} agent.  The agent initiates a discovery operation, after which a bulk regional query is issued.  Histogram(s) of tokens are returned and, based on token frequency, a limited number of decoding operations are executed.  Guidance is subsequently provided back to the MAS.}
  \label{fig:token_agent}
\end{figure*}

A FastAPI-based API is deployed alongside the data store for agentic access.  The API has five general families of endpoints: discover (e.g., identify pertinent frames), local interrogation (e.g., extract time series given lon--lat), regional interrogation (e.g., extract statistics over lon--lat bounding box), similarity (e.g., histogram comparison), and decode.  The API is structured to promote limited data transfer between the client (i.e., where agent lives) and server hosting the data store and API.  In general, agents are prompted to use bulk queries in place of individual point-based queries to ease their processing burden.  A prototypical interaction between \texttt{Token Store} agent and the token store is shown in Figure \ref{fig:token_agent}.  The agent effectively engages in four cascading activities: frame discovery over region, token analysis on discovered frames, limited decoding over flagged frames, and reporting.  In this scenario, a hybrid approach is followed which mixes token-based similarity with decoding.

Regarding token-based similarity analysis, two token grids of the same track are not directly compared to one another.  Rather, their histograms are normalized to probability vectors, and these vectors are used to compute several similarity metrics:
\begin{itemize}
    \item The \texttt{L1} metric is simply one minus half the $L1$ distance between the two distributions.  This metric is fairly general-purpose and used for most analogs.
    \item The \texttt{cos} metric is the cosine similarity of two vectors.  It is used primarily to accentuate similarity of the \textit{shapes} of the distributions, and not necessarily raw counts.  This can be helpful when trying to compare regions with disparate extents of dry grid cells. 
    \item The \texttt{Top-K Jaccard} metric ignores the full histogram and only considers the most frequent codes.  The resulting cropped distributions are used to calculate the Jaccard index, which this metric reports.  It is used to probe top tokens (e.g., when both distributions are dominated by the same few codes) or when \texttt{L1} is washed out by many low-count codes.
\end{itemize}

The API also makes use of a ``patch" decoding strategy wherein, as a screening measure, sub-grids or ``patches" of the $16 \times 16$ token grid are selectively extracted around a location (viz., cell) of interest for decoding.  Decoding these patches is generally 3-4x faster than decoding the full $16 \times 16$ token grid.  It is noteworthy that this sub-sampling breaks the decoder's spatial normalization, so the decoded values are approximate and should be coupled with a full-decode or direct archive retrieval for confirmation.  In the demonstration described below, this patch strategy is used to accelerate token-based screening, after which the archive is sourced directly for final reporting.

To demonstrate the performance of the API and token store, ten queries were issued to the \texttt{Token Store} agent under the SAS modality.  See Table \ref{tab:curated-queries} for the queries that range in complexity from simple retrieval (Q1) to multi-site probing (Q10).  Note that the token-based traversal approach is compared to a non-token-based, conventional traversal approach wherein the archive is sourced directly.  For this comparison, both the token store and archive were hosted on the same machine as the agent client (i.e. co-located), meaning that download overhead was irrelevant.  Of course, download payloads would be relevant in many real-world scenarios where the agent client running on a workstation would have access to archives via S3 or even simple http access (e.g., NOAA's NOMADS), but would not be co-located with the archive itself.  If the archive's chunking strategy were misaligned with the temporal / spatial fidelity of a query, then these payloads could be inordinately large.  Given that most cloud-based archives are already chunked and that re-chunking is largely unfeasible on-the-fly, a user query might fortuitously align with the archive's chunking strategy, meaning that agentic operations would \textit{not} be hindered by inordinately large download payloads and the extra expense in processing them.  Of course, the opposite might be true which would lead to inefficient agentic operations.

For the conventional traversal approach, two versions of the archive were sourced: (1) a map-oriented chunked version representing a ``worst case" for an agent (one full simulation frame per hour), compelling large in-memory processes and (2) a re-chunked version whose chunking strategy is favorable to the queries issued (1 day $\times 4 \times 4$ quadrants).  From Table \ref{tab:q1q10-plane-day4x4-hybrid}, it is clear that map-oriented archive traversal consumes more wall-clock time and requires larger in-memory processing for most queries.  Re-chunked archive traversal rebounds significantly, achieving a 92\% and 83\% reduction in co-located wall-clock time and archive payloads, respectively, on the aggregate.  Nevertheless, the performance of re-chunked-based archival traversal needs to be tempered with the theoretical download payload overhead --- on a 100 Mb/s Ethernet connection, answering the ten queries would require a total of 38 minutes.  Token-based traversal, in turn, exhibits a further 67\% reduction in co-located wall-clock time while requiring $16\times$ less payload (compared to re-chunk-based archive traversal).

\begin{table}[t]
\centering
\small
\caption{Evaluation Queries}
\label{tab:curated-queries}
\begin{tabular}{@{}lp{0.72\linewidth}@{}}
\toprule
ID & Query \\
\midrule
Q1 & What are tide, NTR, and wind at Boston Harbor during the 23~Feb~2026 afternoon wind event? \\
Q2 & What is the maximum wind speed at Boston Harbor over Jan--Jun~2026? \\
Q3 & When does Charleston Harbor reach its peak compound water level (tide\,+\,NTR) in Jan--Jun~2026? \\
Q4 & Where and when is the strongest wind on the Mid-Atlantic shelf (VA--NY) in April~2026? \\
Q5 & Where and when is the highest non-tidal residual on the Florida shelf in March~2026? \\
Q6 & At the NY Bight, when in April~2026 is wind strong (above local p90) while NTR stays flat ($\lvert\mathrm{NTR}\rvert\le 0.15\,\mathrm{m}$)? \\
Q7 & Which hours at Delaware Bay are joint high-tide / high-NTR compound-flood candidates (both $\ge$ local p90) in Jan--Jun~2026? \\
Q8 & How does the Gulf of Mexico NTR footprint evolve during 14--20~Jun~2026? \\
Q9 & What is the coastwide maximum NTR over the East and Gulf coasts during 14--20~Jun~2026? \\
Q10 & Across twelve coastal sites, where/when does high NTR (local p95) coincide with weak wind ($\le 6\,\mathrm{m\,s^{-1}}$) in Jan--Jun~2026? The sites are Galveston, Corpus Christi, New Orleans, Tampa, Miami, Charleston, the Outer Banks, Chesapeake Bay, Delaware Bay, New York Bight, Boston, and Portland (Maine). \\
\bottomrule
\end{tabular}
\end{table}

\begin{table}[t]
\centering
\scriptsize
\caption{Co-located wall-clock time, off-premises wall-clock time (accounting for theoretical 100 Mb/s payload download rate), and estimated download payload for three traversal strategies: direct traversal through a map-oriented archive, direct traversal through a re-chunking of the map-oriented archive, and token-based traversal through the token store.  The re-chunked archive's chunking scheme corresponds to one full day of fields over one quadrant of a $4 \times 4$ split of the grid.  Co-located wall-clock times do \textit{not} reflect network transfer of any kind.  Note that all three traversal strategies produced identical results for all ten queries.}
\label{tab:q1q10-plane-day4x4-hybrid}
\begin{tabular}{@{}l rrr rrr rrr@{}}
\toprule
 & \multicolumn{3}{c}{Map-Oriented Archive} & \multicolumn{3}{c}{Re-Chunked Archive} & \multicolumn{3}{c}{Token Store} \\
\cmidrule(lr){2-4}\cmidrule(lr){5-7}\cmidrule(lr){8-10}
ID & Co-Located & Off-Prem & Payload & Co-Located & Off-Prem & Payload & Co-Located & Off-Prem & Payload \\
 & Wall (s) & Wall (s) & (GiB) & Wall (s) & Wall (s) & (GiB) & Wall (s) & Wall (s) & (GiB) \\
\midrule
Q1 & 0.1 & 1.5 & 0.017 & 0.1 & 2.0 & 0.023 & 0.2 & 2.1 & 0.023 \\
Q2 & 140 & 2239 & 24.433 & 9.1 & 142 & 1.545 & 1.9 & 3.1 & 0.014 \\
Q3 & 100 & 2199 & 24.433 & 5.6 & 138 & 1.545 & 3.6 & 7.1 & 0.040 \\
Q4 & 28 & 376 & 4.054 & 8.3 & 76 & 0.792 & 4.0 & 16 & 0.134 \\
Q5 & 22 & 382 & 4.189 & 11 & 128 & 1.359 & 5.3 & 21 & 0.177 \\
Q6 & 40 & 736 & 8.103 & 2.5 & 48 & 0.530 & 0.8 & 4.2 & 0.040 \\
Q7 & 100 & 2199 & 24.433 & 5.6 & 138 & 1.545 & 3.7 & 8.6 & 0.057 \\
Q8 & 4.7 & 86 & 0.950 & 1.4 & 19 & 0.209 & 1.7 & 11 & 0.107 \\
Q9 & 5.1 & 87 & 0.950 & 45 & 92 & 0.547 & 21 & 50 & 0.346 \\
Q10 & 2849 & 7046 & 48.860 & 175 & 1497 & 15.395 & 45 & 87 & 0.487 \\
\midrule
\textbf{Sum} & \textbf{3288} & \textbf{15350} & \textbf{140.42} & \textbf{264} & \textbf{2281} & \textbf{23.49} & \textbf{87} & \textbf{209} & \textbf{1.425} \\
\bottomrule
\end{tabular}
\end{table}

\section{Integrating MAS and Tokenization Framework}

We now couple the LEG-based MAS with the token-based traversal framework.  Here, we introduce new specialists skilled in token store traversal.  Note that these are highly specialized sub-variants of the \texttt{Token Store} agent.

\begin{itemize}
    \item \texttt{Scout}: Finds plausible locations / periods for which a queried event occurs by sweeping a region and time range rapidly.  It does so with the patch decode strategy discussed earlier.  
    \item \texttt{Tracker}: Follows the queried event along the coastal boxes.  It decodes every coastal box every three hours around the event's peak (timing is received from the \texttt{Scout} agent).  For NTR specifically, it measures how many hours earlier or later each box's rise (or recession) arrives compared to a reference box and subsequently correlates these lags to box location to establish direction and speed of travel.
    \item \texttt{Surveyor}: It decodes un-patched grids for extracting data for validation.  It does not read raw tokens.
    \item \texttt{Skeptic}: Checks whether the salient conclusions from the workflow can be trusted.  It computes errors from observations when available.
\end{itemize}

\begin{figure*}[t]
  \centering
  \includegraphics[width=0.8\textwidth]{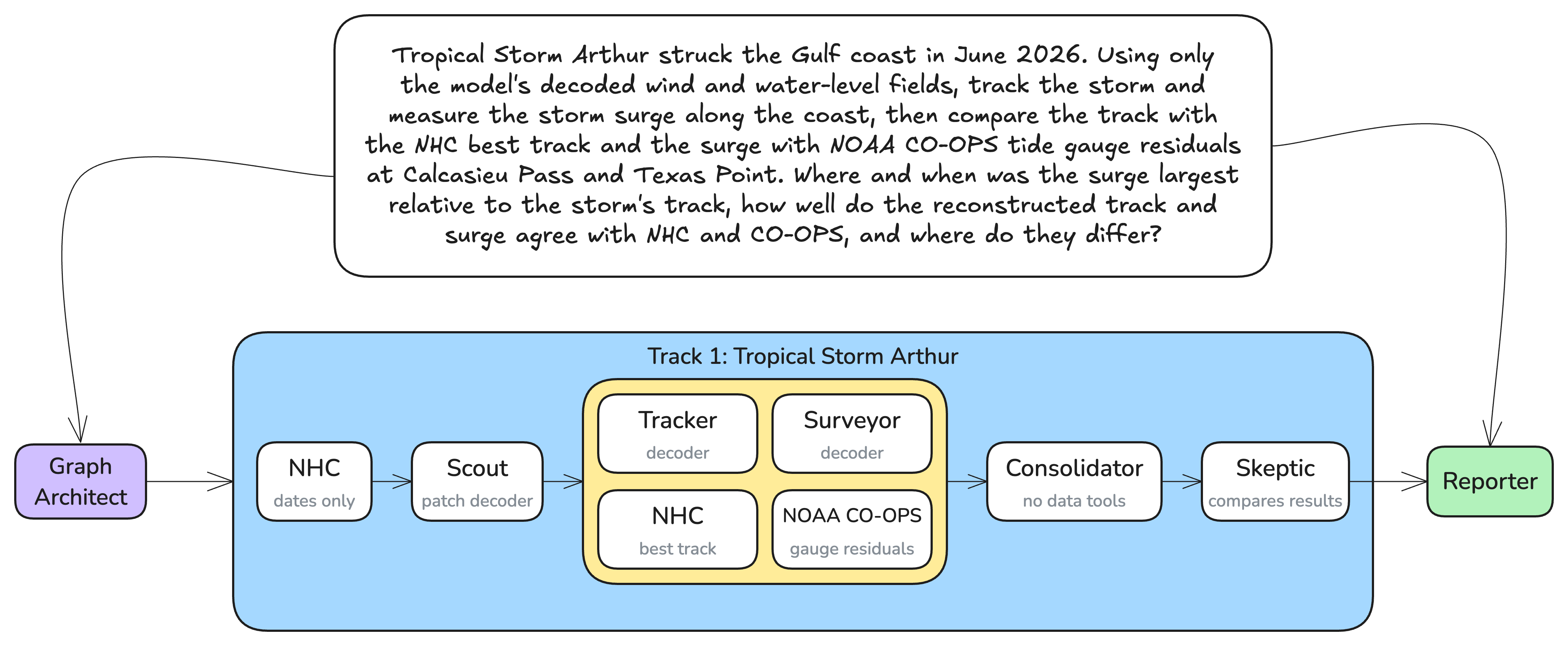}
  \caption{LEG with token store specialists.}
  \label{fig:mas_token}
\end{figure*}

For demonstration, we consider a query that is significantly more open-ended than the queries considered in the previous section as it involves tracking and validation.  The query, which is shown in Figure \ref{fig:mas_token}, involves tracking Tropical Storm Arthur (2026) through the token store and comparing the reconstruction to the best track from the \texttt{NHC} agent.  Additionally, it invokes the \texttt{NOAA CO-OPS} agent for validation at select stations.  Here, the \texttt{Graph Architect} agent divides the workflow among several agents in a one-track LEG, Figure \ref{fig:mas_token}.  A truncated version of the \texttt{Reporter} agent's output is shown below:

\begin{verbatim}
MAS REPORTER OUTPUT (truncated)
================================
Query:
  Tropical Storm Arthur struck the Gulf coast in June 2026. Using only
  the model's decoded wind and water-level fields, track the storm and
  measure the storm surge along the coast, then compare the track with
  the NHC best track and the surge with NOAA CO-OPS tide gauge residuals
  at Calcasieu Pass and Texas Point. Where and when was the surge
  largest relative to the storm's track, how well do the reconstructed
  track and surge agree with NHC and CO-OPS, and where do they differ?

Short answer
------------
  Track: median error 62 km (mean 68, max 149) against the NHC best
  track over 10 comparisons at 3-hourly times, 16 Jun 18:00Z to 17 Jun
  21:00Z; on average the reconstructed centre is 48 km ahead of NHC
  along its motion and 26 km to the right.
  Surge: largest decoded surge 0.63 m near Sabine Pass at 17 Jun 17:30Z.
  Gauges: 2 CO-OPS gauges, median peak-time difference 5.0 h, median
  correlation 0.87, median decoded/observed peak rise 0.99.
  Verification: PASS (7 of 7 checks passed).

Largest decoded surge peaks
---------------------------
  1. Sabine Pass   0.63 m  17 Jun 17:30Z
  2. Cameron       0.52 m  17 Jun 19:30Z
  3. Eugene Island 0.50 m  18 Jun 05:30Z*
  (* after the last NHC fix)

Decoded vs CO-OPS (peak rise above pre-storm level)
---------------------------------------------------
  Texas Point    obs 0.68 m 17 Jun 16:00Z  dec 0.56 m 17 Jun 21:00Z
  Calcasieu Pass obs 0.54 m 17 Jun 23:00Z  dec 0.62 m 18 Jun 08:00Z

\end{verbatim}

Arthur (2026) was relatively disorganized, meaning that reconstructing its track from the GFS wind field is an inherently error-prone process.  The relative difference between GFS-based and token-based reconstructions is relatively small, Table \ref{tab:arthur-gfs-vs-token}, confirmation that token-based traversal faithfully preserves the GFS field.

Of course, this is one sample MAS-token integration.  We view MAS as a means not necessarily to accelerate token-based archive traversal, but to support agentic hypothesis formulation.  The more specialists available, the more data that can be sourced (be it observation or modeled).  This means that richer hypotheses can be postulated and tested more efficiently.  

\begin{table}[t]
\centering
\scriptsize
\caption{Position of Tropical Storm Arthur (2026) reconstructed from the raw GFS wind field and from the decoded wind field.  Both reconstructions use the same vorticity-tracking method.}
\label{tab:arthur-gfs-vs-token}
\begin{tabular}{@{}l rr rrr rrr r@{}}
\toprule
 & \multicolumn{2}{c}{NHC} & \multicolumn{3}{c}{GFS Wind} & \multicolumn{3}{c}{Decoded Tokens} & Token vs GFS \\
\cmidrule(lr){2-3}\cmidrule(lr){4-6}\cmidrule(lr){7-9}
Time & Lat. & Lon. & Lat. & Lon. & Error & Lat. & Lon. & Error & Distance \\
(UTC) & ($^{\circ}$N) & ($^{\circ}$W) & ($^{\circ}$N) & ($^{\circ}$W) & (km) & ($^{\circ}$N) & ($^{\circ}$W) & (km) & (km) \\
\midrule
16~Jun 18:00Z & 27.10 & 97.80 & 27.47 & 96.68 & 118 & 27.70 & 97.02 & 102 & 42 \\
16~Jun 21:00Z & 27.25 & 97.60 & 28.17 & 96.80 & 130 & 27.60 & 96.28 & 136 & 82 \\
17~Jun 00:00Z & 27.40 & 97.40 & 28.06 & 96.21 & 138 & 28.12 & 96.13 & 149 & 11 \\
17~Jun 03:00Z & 27.55 & 97.25 & 27.94 & 97.15 & 45 & 27.55 & 97.09 & 16 & 44 \\
17~Jun 06:00Z & 27.70 & 97.10 & 27.71 & 97.03 & 7 & 27.67 & 97.00 & 10 & 5 \\
17~Jun 09:00Z & 28.05 & 96.60 & 27.94 & 96.80 & 23 & 27.85 & 96.78 & 28 & 10 \\
17~Jun 12:00Z & 28.40 & 96.10 & 28.41 & 95.98 & 12 & 28.34 & 95.92 & 19 & 9 \\
17~Jun 15:00Z & 28.60 & 96.10 & 28.41 & 95.86 & 32 & 28.46 & 95.63 & 48 & 23 \\
17~Jun 18:00Z & 28.80 & 96.10 & 28.64 & 95.04 & 105 & 28.70 & 95.13 & 95 & 11 \\
17~Jun 21:00Z & 29.10 & 95.45 & 28.76 & 95.74 & 47 & 29.12 & 94.66 & 76 & 112 \\
\midrule
\textbf{Median} & & & & & \textbf{46} & & & \textbf{62} & \textbf{17} \\
\textbf{Mean} & & & & & \textbf{66} & & & \textbf{68} & \textbf{35} \\
\textbf{Maximum} & & & & & \textbf{138} & & & \textbf{149} & \textbf{112} \\
\bottomrule
\end{tabular}
\end{table}

\section{Conclusions}
\label{sec:conclusions}
Agentic operations in hydrodynamics are currently limited by sheer scale.  For example, coastal hydrodynamic models tend to preserve high fidelity in both the spatial and temporal dimensions; however, agents cannot simply ingest this data without some penalty, and this manifests as context saturation and associated hallucinations.  Multi-agent systems (MAS) can reduce this context burden by spreading labor over multiple agents, but MAS alone does not sufficiently address the saturation problem as archives themselves are not optimized for agentic operations.  Agents continuously refine their workplans and operate fairly non-deterministically, and this means that they might need to source data from hydrodynamic-derived archives in unanticipated ways.  This requires some level of flexibility on the part of the archive itself, one that facilitates exploration.  In response, we have proposed a token-based traversal layer for hydrodynamic datasets to support rapid agentic reasoning.

Tokens are the codebook indices produced by vector quantization.  Herein, our tokens were produced by vector quantized variational autoencoders (VQ-VAEs).  Grids of these tokens discretized the US Atlantic and Gulf Coasts, describing tide level, non-tidal residual, and surface winds.  They were populated in a database called the \textit{token store} with both time and location-based entry points.  Agents traversing the store were able to decode full grids of tokens on-the-fly and engage in rapid histogram-based similarity analyses without decoding.  We compared this token-based traversal strategy to conventional archive-based techniques on ten open-ended (i.e., under-specified) queries and demonstrated significantly lower run-times and, for off-premise applications, download payloads.  Specifically, compared to traversing a strategically chunked archive, token-based traversal achieved a $3\times$ faster run-time and $16\times$ reduced download payload.  We then demonstrated token-based traversal within a MAS, showcasing its ability to track trends through token signals.

Operational models in both the atmospheric and oceanic sectors archive their data in ways optimized for storage and not necessarily access.  Placing a thin layer on top of an archive in the form of a token store can facilitate traversal and ultimately agentic reasoning by reducing download payloads, relieving pressure off of context windows, and accelerating hypothesis-driven agentic activities.  Ultimately, the traversal scheme must be adapted to the dataset.  When approaching token-based traversal from the outset, a developer must reconcile the physical dimensions (viz., time, lon, lat, and $z$ if relevant) to the latent space.  While the token store introduced herein, filled with its time-decorated boxes dotting the US Gulf and Atlantic Coasts, is one such attempt at this reconciliation, we emphasize that co-locating the grids of tokens into databases \textit{specifically} for agents is relatively unexplored and further study could reveal greater gains than those described herein.


\section*{Acknowledgments}
We acknowledge support from the Wellcome Trust under the project ``FloDisMod: A Framework for Flood and Disease Modeling'' (226092/Z/22/Z) and by a subcontract from Los Alamos National Laboratory under Subcontract C5223 and Master Agreement BA\# 588340.  We also acknowledge the work of Dr. Mansur Jisan of NOAA in creating OceanMCP.  A handful of tools in OceanMCP were adapted for use in some of the specialist agents herein.

\bibliographystyle{unsrt}  
\bibliography{references}

\newpage

\begin{appendices}
\counterwithin{table}{section}

\section{Performance Assessment of the MAS Prototype}
\label{app:assessment}

This appendix evaluates the orchestration layer of the MAS prototype in Section~\ref{sec:prototype}, that is, how the \texttt{Graph Architect} plans a LEG, how the specialists retrieve data within it, and how the \texttt{Consolidator} and \texttt{Reporter} turn that data into an answer. The queries rely on four specialists (\texttt{NHC}, \texttt{NOAA CO-OPS}, \texttt{USGS}, and \texttt{FEMA}). Each one draws on an authoritative federal source, so answers can be scored against established ground truth. These sources also return compact responses to clearly specified queries, which keeps the focus on orchestration rather than on handling large datasets. All results are reported in Table~\ref{tab:assessment}.

\subsection{Setup and Metrics}

The benchmark has 37 queries in six complexity categories, listed in Tables~\ref{tab:app_single} through~\ref{tab:app_parallel}. They range from single specialist lookups (\texttt{NHC} storm categories and counts, \texttt{FEMA} flood zones), to linear pipelines in which \texttt{NHC} precedes \texttt{NOAA CO-OPS} or \texttt{NOAA CO-OPS}{+}\texttt{USGS}, to parallel track queries with two or three independent subquestions. Ground truth comes from eight NHC Tropical Cyclone Reports (AL092008, AL182012, AL122005, AL092017, AL142018, AL092022, AL132003, AL092004), NOAA Technical Reports NOS CO-OPS~079 and~067, the NOAA CO-OPS Hurricane Irma Water Level Report, NHC Atlantic Basin Storm Totals, and the FEMA NFHL MapServer. 

For the ablations and the cost analysis, we use a seven query subset that covers all six categories and is marked with an asterisk in Tables~\ref{tab:app_single} through~\ref{tab:app_parallel}. Two of these queries come from the NHC$\to$NOAA category because they stress the system in different ways. L01 involves a well instrumented station with continuous data, while L03 involves a gauge that went offline during Hurricane Katrina, which makes it the hardest surge query in the set. Every experiment used Claude Sonnet~4.6 as the backbone LLM for both the specialist and general-purpose agents. For the ablations, we did not modify any model code. Each variant intercepts function calls at runtime to bypass the targeted component while the rest of the pipeline runs as usual. 

Each query is scored with four metrics. 

\begin{itemize}
    \item \textbf{Factual Precision} measures how closely the reported value matches an official reference value.  For surge queries, the reference $y$ is the documented peak surge (in meters) from the relevant NHC Tropical Cyclone Report or NOAA CO-OPS Technical Report, and the system's closest extracted value $\hat{y}$ receives a score of $\max(0,\, 1 - |\hat{y} - y|/y)$.  The same formula applies to season count queries with integer references.  For storm category queries, the score is binary (correct Saffir-Simpson category or not).  For flood zone queries, it is the fraction of expected FEMA zone codes mentioned.  For queries with several parts spanning parallel tracks, the score is averaged across subquestions.

    \item \textbf{Topology Selection} is 1 if the \texttt{Graph Architect} chooses the correct LEG topology (\texttt{linear} or \texttt{parallel\_tracks}) and 0 otherwise.

    \item \textbf{Agent F1} is the harmonic mean of precision and recall for specialist selection, where precision penalizes superfluous agents and recall penalizes missing ones.

    \item \textbf{Source Attribution} measures how well the pipeline cites its data sources. For surge queries, it has four equally weighted components (station ID, vertical datum, temporal reference, and authoritative source name). These are evaluated over the combined text of the \texttt{Consolidator}, \texttt{Cross-Track Merge}, and \texttt{Reporter} outputs, so details kept in intermediate stages but condensed by the \texttt{Reporter} still receive credit. Count, category, and flood zone queries use analogous scores with two components, such as database name and year for count queries, or source name and zone code for flood zone queries. 

\end{itemize}

A query's overall score is the unweighted mean of these four metrics. 

\begin{table}[p]
\caption{\textbf{Performance assessment of the MAS prototype.}  In panel (b), $\Delta$ is the change from the full system.}
\label{tab:assessment}
\centering

\vspace{0.3em}
{\footnotesize\textbf{(a) Benchmark results by query category} ($N{=}37$, values in \%)}\\[3pt]
\footnotesize
\setlength{\tabcolsep}{4pt}
\begin{tabular}{l cc cc cc c}
\toprule
& \multicolumn{2}{c}{Single specialist} & \multicolumn{2}{c}{Linear pipeline} & \multicolumn{2}{c}{Parallel tracks} & \\
\cmidrule(lr){2-3}\cmidrule(lr){4-5}\cmidrule(lr){6-7}
Metric & NHC & FEMA/NOAA & NHC$\to$NOAA & NHC$\to$N{+}U & 2 tracks & 3 tracks & Overall \\
& ($n{=}7$) & ($n{=}5$) & ($n{=}8$) & ($n{=}5$) & ($n{=}7$) & ($n{=}5$) & ($N{=}37$) \\
\midrule
Factual precision  & 99.5  & 95.0  & 90.1  & 82.2  & 94.9  & 95.9  & 93.2 \\
Topology selection & 100.0 & 100.0 & 100.0 & 100.0 & 100.0 & 100.0 & 100.0 \\
Agent F1           & 100.0 & 100.0 & 80.0  & 88.0  & 95.1  & 97.1  & 92.7 \\
Source attribution & 100.0 & 100.0 & 90.6  & 70.0  & 78.6  & 90.0  & 88.5 \\
\midrule
Overall score      & 99.9  & 98.8  & 90.2  & 85.0  & 92.2  & 95.8  & \textbf{93.6} \\
Avg.\ latency (s)  & 18    & 25    & 89    & 68    & 73    & 93    & 62 \\
\bottomrule
\multicolumn{8}{l}{\scriptsize Pass rate (overall score $\geq$ 50\%) 37/37, with 0 execution errors and a total run time of 38.0\,min.}
\end{tabular}

\vspace{1.0em}

{\footnotesize\textbf{(b) Ablations} (subset of seven queries, values in \%)}\\[3pt]
\footnotesize
\setlength{\tabcolsep}{4pt}
\begin{tabular}{l c cc cc cc}
\toprule
& Full system & \multicolumn{2}{c}{Fixed graph} & \multicolumn{2}{c}{No consolidation} & \multicolumn{2}{c}{No reporter} \\
\cmidrule(lr){3-4}\cmidrule(lr){5-6}\cmidrule(lr){7-8}
Metric & Score & Score & $\Delta$ & Score & $\Delta$ & Score & $\Delta$ \\
\midrule
Factual precision  & 91.9  & 63.2 & $-$28.7 & 87.3  & $-$4.6 & 81.1  & $-$10.8 \\
Topology selection & 100.0 & 71.4 & $-$28.6 & 100.0 & 0.0    & 100.0 & 0.0 \\
Agent F1           & 87.3  & 67.2 & $-$20.1 & 87.3  & 0.0    & 87.3  & 0.0 \\
Source attribution & 84.5  & 59.5 & $-$25.0 & 93.5  & $+$8.9 & 88.1  & $+$3.6 \\
\midrule
Avg.\ latency (s)  & 75    & 85   & $+$10   & 65    & $-$10  & 74    & $-$1 \\
\bottomrule
\multicolumn{8}{l}{\scriptsize The fixed graph routes every query through \texttt{[NHC]}$\to$\texttt{[NOAA CO-OPS, USGS]}$\to$\texttt{[FEMA]}.}
\end{tabular}

\vspace{1.0em}

{\footnotesize\textbf{(c) Complexity scaling} (independent subquestions per query)}\\[3pt]
\footnotesize
\setlength{\tabcolsep}{6pt}
\begin{tabular}{l ccccc}
\toprule
Subquestions & 1 & 2 & 3 & 4 & 5 \\
\midrule
$n$                & 25   & 7    & 5    & 1    & 1 \\
Overall score (\%) & 93.6 & 92.2 & 95.8 & 90.6 & 95.0 \\
Latency (s)        & 52   & 73   & 93   & 137  & 152 \\
\bottomrule
\end{tabular}

\vspace{1.0em}

{\footnotesize\textbf{(d) Robustness}}\\[3pt]
\footnotesize
\setlength{\tabcolsep}{4pt}
\begin{tabular}{lc lc lc}
\toprule
\multicolumn{2}{c}{Paraphrase (4 queries $\times$ 3)} & \multicolumn{2}{c}{Adversarial (5 queries)} & \multicolumn{2}{c}{Fault tolerance (3 outages)} \\
\cmidrule(lr){1-2}\cmidrule(lr){3-4}\cmidrule(lr){5-6}
Same topology       & 4/4 & Premise detected     & 4/5 & Partial answer     & 3/3 \\
Same specialists    & 2/4 & No fabricated data   & 5/5 & Missing data noted & 2/3 \\
Score $\sigma$ (\%) & 1.2 & Appropriate response & 4/5 & No crash           & 3/3 \\
\bottomrule
\multicolumn{6}{l}{\scriptsize Specialist sets differed for L01 and P01 (score $\sigma$ of 2.7\% and 2.0\%) and matched for S02 and O01 ($\sigma$ of 0.0\%).}
\end{tabular}

\vspace{1.0em}

\begin{minipage}[t]{0.56\linewidth}
\centering
{\footnotesize\textbf{(e) Cost per query} (subset of seven queries)}\\[3pt]
\footnotesize
\setlength{\tabcolsep}{4pt}
\begin{tabular}{ll rrr}
\toprule
ID & Category & Tokens & Cost (\$) & Time (s) \\
\midrule
S02 & Single NHC     & 16,546  & 0.06 & 21 \\
O01 & Single FEMA    & 14,910  & 0.06 & 43 \\
M01 & NHC$\to$N{+}U  & 55,997  & 0.19 & 58 \\
L01 & NHC$\to$NOAA   & 96,959  & 0.33 & 78 \\
L03 & NHC$\to$NOAA   & 136,460 & 0.45 & 88 \\
P01 & 2 tracks       & 103,097 & 0.38 & 118 \\
C01 & 3 tracks       & 103,525 & 0.38 & 119 \\
\midrule
\multicolumn{2}{l}{Average} & 75,356 & 0.27 & 75 \\
\bottomrule
\multicolumn{5}{l}{\scriptsize Tokens are summed over all pipeline stages.} \\
\multicolumn{5}{l}{\scriptsize Extrapolated to all 37 queries, about \$9.85 and 2.8M tokens.}
\end{tabular}
\end{minipage}%
\hfill
\begin{minipage}[t]{0.42\linewidth}
\centering
{\footnotesize\textbf{(f) Tokens by pipeline stage}}\\[3pt]
\footnotesize
\setlength{\tabcolsep}{4pt}
\begin{tabular}{l rrr}
\toprule
Stage & Input & Output & Share (\%) \\
\midrule
Architect    & 21,533  & 1,572  & 4.4 \\
Specialist   & 468,643 & 17,515 & 92.2 \\
Consolidator & 4,697   & 1,869  & 1.2 \\
Reporter     & 9,227   & 2,438  & 2.2 \\
\midrule
Total        & 504,100 & 23,394 & 100.0 \\
\bottomrule
\end{tabular}
\end{minipage}

\end{table}

\subsection{Results}

Table~\ref{tab:assessment}(a) breaks the benchmark results down by category. The prototype answered all 37 queries without an execution error and scored 93.6\% overall. Factual precision was highest for single specialist \texttt{NHC} queries (99.5\%), which are direct database lookups. It was lowest for the NHC$\to$N{+}U category (82.2\%), where three specialists have to reconcile gauge observations with surveyed high water marks from two federal sources that use different vertical datums. The \texttt{Graph Architect} chose the correct topology for every query, which suggests the natural language routing heuristics in Section~\ref{sec:prototype} are enough to guide that decision. Most of the loss in agent F1 comes from one of those heuristics. For surge queries that do not name a source, it adds the \texttt{USGS} specialist together with \texttt{NOAA CO-OPS}, which gives more complete data at some cost to selection precision. Source attribution was 100\% for single specialist categories and between 70\% and 91\% for surge queries that involve several specialists, depending on whether the \texttt{Reporter} kept station identifiers and datum references in its prose. 

The ablations in Table~\ref{tab:assessment}(b) each disable a single component and leave the rest of the pipeline alone. Replacing the \texttt{Graph Architect} with a static LEG, which mimics a rigid and programmatic workflow, did the most damage. Factual precision fell by 28.7 percentage points (pp), partly because the static graph cannot split a query with several parts into parallel tracks, so agents end up juggling unrelated data at once. Irrelevant specialists also add noise to the context passed to the \texttt{Reporter}, as with \texttt{FEMA} on a pure surge query. Swapping consolidation within each layer and across tracks for raw text concatenation lowered factual precision by 4.6~pp, and skipping the \texttt{Reporter} so that the \texttt{Consolidator}'s internal brief became the final answer lowered it by 10.8~pp. Source attribution went up in those two cases, since raw specialist output carries more station IDs, datums, and timestamps than synthesized prose. We do not count this as a gain, because the user receives verbose tool output instead of a coherent report.

Panels (c) and (d) cover the stress tests, whose queries are listed in  Appendix~\ref{app:stress_queries}. As queries grew from one to five independent subquestions, scores stayed between 90.6\% and 95.8\%, and latency rose by about 25\,s per added track, which is consistent with parallel specialist execution plus a fixed merge across tracks. We also rephrased each of four queries in two additional ways, and the LEG topology never changed. The specialist set was less stable, because shorter phrasings sometimes failed to trigger the heuristic that adds a complementary specialist. Even so, the factual score barely changed across phrasings. Its standard deviation, averaged over the four queries, was 1.2\%.

We checked each adversarial response on whether it flagged the problematic premise, avoided fabricating data, and matched the expected behavior in Table~\ref{tab:app_adversarial}. The system never fabricated data, and it flagged the premise and responded as expected in four out of five cases. For Denver, it gave a minimal response instead of an explicit rejection. When told that Ike was a Category~5 storm, it pointed out that ``Ike made landfall as a Category~2\ldots'' and then answered the underlying surge question. 

Finally, we simulated an outage of one specialist at a time by forcing all of its tool calls to return a structured error message. We did this for \texttt{NOAA CO-OPS}, \texttt{USGS}, and \texttt{FEMA}. A run passed if it finished without crashing and still gave a substantive partial answer from the remaining sources. We also noted whether the response mentioned the missing data. All three runs passed, and two of the three responses mentioned the missing source. The \texttt{NOAA CO-OPS} run did not, but its answer was still built from NHC track data and USGS high water marks. Each specialist runs with its own set of allowed tools and its own context, so one failing agent does not affect the others.

Panels (e) and (f) report token use and cost at published API rates of \$3 per million input tokens and \$15 per million output tokens. Single specialist queries cost \$0.06, while queries with several specialists or parallel tracks cost \$0.19 to \$0.45. The most expensive was L03, where a gap in the gauge record sent the \texttt{NOAA CO-OPS} specialist through extra tool rounds looking for alternative stations and date ranges. Specialists used 92.2\% of all tokens, so planning, merging, and reporting add little on top of the actual data retrieval.

\section{Evaluation Queries}
\label{app:queries}

\subsection{Benchmark Queries}
\label{appendix:queries}
Tables~\ref{tab:app_single} through~\ref{tab:app_parallel} list the 37 benchmark queries and their ground truth sources. 

\begin{table*}[!htbp]
\centering
\caption{Queries handled by a single specialist ($n=12$).  Asterisks mark the ablation subset.}
\label{tab:app_single}
\vspace{2pt}
\small
\querytablestyle
\begin{tabular}{llp{7cm}p{5cm}}
\toprule
\rowcolor{slate!20}
\textbf{ID} & \textbf{Category} & \textbf{Query} & \textbf{Ground Truth Source} \\
\midrule
S01 & NHC & How many named storms were in the 2005 Atlantic hurricane season? & 28 named (NHC Storm Totals) \\
S02$^{*}$ & NHC & What category was Hurricane Michael when it made landfall in Florida in 2018? & Cat~5, 140\,kt (TCR AL142018) \\
S03 & NHC & What category was Hurricane Harvey at landfall in Texas in 2017? & Cat~4, 115\,kt (TCR AL092017) \\
S04 & NHC & What was the peak intensity of Hurricane Katrina in 2005? & Cat~5, 150\,kt (TCR AL122005) \\
S05 & NHC & How many hurricanes occurred in the 2017 Atlantic season? & 10 hurricanes (NHC Storm Totals) \\
S06 & NHC & How many named storms were in the 2020 Atlantic hurricane season? & 30 named (NHC Storm Totals) \\
S07 & NHC & What category was Hurricane Ike when it made landfall in Texas in 2008? & Cat~2, 95\,kt (TCR AL092008) \\
\midrule
O01$^{*}$ & FEMA & What are the FEMA flood zones for Miami Beach, Florida? & AE, VE, SFHA (NFHL MapServer) \\
O02 & FEMA & What is the FEMA flood zone designation for Tampa, Florida? & AE, SFHA (NFHL MapServer) \\
O03 & NOAA CO-OPS & What are the current water level conditions at San Francisco tide station? & Live data, stn 9414290 (CO-OPS API) \\
O04 & FEMA & What is the FEMA flood zone for Galveston, Texas? & VE, AE, SFHA (NFHL MapServer) \\
O05 & FEMA & What are the FEMA flood zones for Charleston, South Carolina? & AE, VE, SFHA (NFHL MapServer) \\
\bottomrule
\end{tabular}
\end{table*}

\begin{table*}[!htbp]
\centering
\caption{Linear pipeline queries ($n=13$).  NHC$\to$NOAA denotes a LEG with two layers, and NHC$\to$N{+}U denotes \texttt{NHC} followed by \texttt{NOAA CO-OPS} and \texttt{USGS} in parallel.  Asterisks mark the ablation subset.}
\label{tab:app_linear}
\vspace{2pt}
\small
\querytablestyle
\begin{tabular}{llp{7cm}p{5cm}}
\toprule
\rowcolor{slate!20}
\textbf{ID} & \textbf{Category} & \textbf{Query} & \textbf{Ground Truth Source} \\
\midrule
L01$^{*}$ & NHC$\to$NOAA & What was the storm surge at Galveston during Hurricane Ike in 2008? & 2.44\,m, stn 8771450 (CO-OPS~079) \\
L02 & NHC$\to$NOAA & What was the peak storm surge at The Battery, New York during Hurricane Sandy in 2012? & 2.81\,m, stn 8518750 (TCR AL182012) \\
L03$^{*}$ & NHC$\to$NOAA & What was the storm surge at Grand Isle during Hurricane Katrina in 2005? & ${\sim}$2.5\,m, stn 8761724, gauge gap (CO-OPS~067) \\
L04 & NHC$\to$NOAA & What was the storm surge at Key West during Hurricane Irma in 2017? & 0.98\,m, stn 8724580 (CO-OPS Irma Report) \\
L05 & NHC$\to$NOAA & What was the observed storm surge at Sabine Pass during Hurricane Ike? & 3.90\,m (12.79\,ft), Sabine Pass North (TCR AL092008) \\
L06 & NHC$\to$NOAA & What was the storm surge at Dauphin Island during Hurricane Katrina? & 1.64\,m, stn 8735180 (CO-OPS~067) \\
L07 & NHC$\to$NOAA & What was the storm surge at Virginia Key, Miami during Hurricane Irma? & 0.61\,m, stn 8723214 (CO-OPS Irma Report) \\
L08 & NHC$\to$NOAA & What was the observed storm surge at Sewells Point, Virginia during Hurricane Isabel in 2003? & 1.46\,m, stn 8638610 (TCR AL132003) \\
\midrule
M01$^{*}$ & NHC$\to$N{+}U & What was the observed storm surge and high water marks near Fort Myers during Hurricane Ian in 2022? & Gauge 7.26\,ft MHHW, HWMs 10 to 15\,ft (TCR AL092022) \\
M02 & NHC$\to$N{+}U & What were the storm surge observations and surveyed peak water levels at Galveston during Hurricane Ike? & CO-OPS ${\sim}$2.44\,m, HWMs 4.6 to 6.1\,m (CO-OPS~079) \\
M03 & NHC$\to$N{+}U & What was the observed storm surge and high water marks near Naples during Hurricane Ian? & Naples gauge, USGS ${>}$9\,ft MHHW (TCR AL092022) \\
M04 & NHC$\to$N{+}U & What were the observed surge and surveyed peak water levels near Pensacola during Hurricane Ivan in 2004? & ${\sim}$3 to 5\,m east of eye (TCR AL092004) \\
M05 & NHC$\to$N{+}U & What were the observed surge and high water marks during Hurricane Sandy at Kings Point, New York? & 3.56\,m, stn 8516945 (TCR AL182012) \\
\bottomrule
\end{tabular}
\end{table*}

\begin{table*}[!htbp]
\centering
\caption{Queries with parallel tracks ($n=12$).  2T and 3T denote the expected number of independent parallel tracks.  Asterisks mark the ablation subset.}
\label{tab:app_parallel}
\vspace{2pt}
\small
\querytablestyle
\begin{tabular}{llp{9.3cm}l}
\toprule
\rowcolor{slate!20}
\textbf{ID} & \textbf{Tracks} & \textbf{Query} & \textbf{Sources} \\
\midrule
P01$^{*}$ & 2T & What was the storm surge at Galveston during Hurricane Ike in 2008, and what are the FEMA flood zones for Miami Beach? & CO-OPS~079 + NFHL \\
P02 & 2T & How many named storms were in the 2005 season, and what was the peak surge at The Battery during Hurricane Sandy? & NHC Totals + TCR \\
P03 & 2T & What category was Hurricane Harvey at landfall, and what are the FEMA flood zones near Corpus Christi, Texas? & TCR + NFHL \\
P04 & 2T & What was the storm surge at Key West during Irma in 2017, and how many hurricanes were in the 2017 season? & CO-OPS Irma + NHC \\
P05 & 2T & What was the storm surge at Dauphin Island during Katrina, and what are the FEMA flood zones for Galveston? & CO-OPS~067 + NFHL \\
P06 & 2T & What category was Hurricane Ike at landfall, and what was the storm surge at The Battery during Sandy? & TCRs \\
P07 & 2T & How many named storms were in the 2020 season, and what are the FEMA flood zones for Charleston? & NHC Totals + NFHL \\
\midrule
C01$^{*}$ & 3T & What was the observed storm surge at Galveston during Hurricane Ike in 2008, and how many named storms were in the 2005 Atlantic season, and what are the FEMA flood zones for Tampa? & CO-OPS~079 + NHC + NFHL \\
C02 & 3T & What was the storm surge at The Battery during Sandy, and what category was Hurricane Michael at landfall, and what is the FEMA flood zone for Miami Beach? & TCRs + NFHL \\
C03 & 3T & What was the surge at Key West during Irma, and how many hurricanes were in the 2017 season, and what are the FEMA flood zones for Galveston? & CO-OPS Irma + NHC + NFHL \\
C04 & 3T & What category was Harvey at landfall in 2017, and what was the storm surge at Dauphin Island during Katrina, and what are the FEMA flood zones for Tampa? & TCRs + CO-OPS~067 + NFHL \\
C05 & 3T & What was the surge at Galveston during Ike, and what category was Hurricane Katrina at peak intensity, and what are the FEMA flood zones for Charleston? & CO-OPS~079 + TCR + NFHL \\
\bottomrule
\end{tabular}
\end{table*}

\subsection{Stress Test Queries}
\label{app:stress_queries}

Table~\ref{tab:app_paraphrase} lists the paraphrase variants and Table~\ref{tab:app_adversarial} the adversarial queries. The two longer queries added for the complexity scaling test and the query used in all three outage scenarios are listed in Table~\ref{tab:app_stress_extra}.

\begin{table*}[!htbp]
\centering
\caption{Paraphrase variants.  Variant~(a) is the original benchmark phrasing, while (b) and (c) test casual and shorter phrasings.}
\label{tab:app_paraphrase}
\vspace{2pt}
\small
\querytablestyle
\begin{tabular}{llp{12.5cm}}
\toprule
\rowcolor{slate!20}
\textbf{Group} & \textbf{Var.} & \textbf{Query} \\
\midrule
L01 & a & What was the storm surge at Galveston during Hurricane Ike in 2008? \\
    & b & How high did the water surge get in Galveston, TX when Ike hit in 2008? \\
    & c & Ike 2008 --- observed surge height at Galveston tide gauge? \\
\midrule
S02 & a & What category was Hurricane Michael when it made landfall in Florida in 2018? \\
    & b & Michael 2018 --- what Saffir--Simpson category at Florida landfall? \\
    & c & How strong was Hurricane Michael when it struck the Florida coast in 2018? \\
\midrule
O01 & a & What are the FEMA flood zones for Miami Beach, Florida? \\
    & b & What FEMA NFHL flood zone designation applies to Miami Beach? \\
    & c & Is Miami Beach in a special flood hazard area?  What zones does FEMA show? \\
\midrule
P01 & a & What was the storm surge at Galveston during Hurricane Ike in 2008, and what are the FEMA flood zones for Miami Beach? \\
    & b & Two questions. (1) How high was the surge in Galveston from Ike 2008, and (2) what FEMA flood zones cover Miami Beach? \\
    & c & I need the Ike 2008 storm surge data for Galveston and also the FEMA flood zone map for Miami Beach. \\
\bottomrule
\end{tabular}
\end{table*}

\begin{table*}[!htbp]
\centering
\caption{Adversarial queries and expected behavior.}
\label{tab:app_adversarial}
\vspace{2pt}
\small
\querytablestyle
\begin{tabular}{llp{7cm}p{4.5cm}}
\toprule
\rowcolor{slate!20}
\textbf{ID} & \textbf{Type} & \textbf{Query} & \textbf{Expected Behavior} \\
\midrule
ADV1 & Wrong location & What was the storm surge at Galveston during Hurricane Sandy in 2012? & Clarify Sandy did not affect Galveston \\
ADV2 & Inland location & What was the storm surge in Denver, Colorado? & Reject, as there is no coastal data \\
ADV3 & Nonexistent storm & What was the surge and category of Hurricane Bob in 2008? & Report storm not found in HURDAT2 \\
ADV4 & Wrong category & What was the Category~5 storm surge at Galveston during Hurricane Ike in 2008? & Correct to Category~2 and still provide surge \\
ADV5 & Inland FEMA & What are the FEMA coastal flood zones for Omaha, Nebraska? & Note riverine only, not coastal \\
\bottomrule
\end{tabular}
\end{table*}

\begin{table*}[!htbp]
\centering
\caption{Additional stress test queries}
\label{tab:app_stress_extra}
\vspace{2pt}
\small
\querytablestyle
\begin{tabular}{p{4.5cm}p{11cm}}
\toprule
\rowcolor{slate!20}
\textbf{Test} & \textbf{Query} \\
\midrule
Complexity scaling, four tracks & What was the storm surge at Galveston during Ike in 2008, and what category was Hurricane Michael at landfall, and what are the FEMA flood zones for Miami Beach, and how many named storms were in the 2020 season? \\
Complexity scaling, five tracks & What was the storm surge at The Battery during Sandy, and what category was Hurricane Harvey at landfall, and what are the FEMA flood zones for Tampa, and how many hurricanes were in the 2017 season, and what was the surge at Key West during Irma? \\
Fault tolerance, same query for all three outages & What was the observed storm surge and high water marks at Galveston during Hurricane Ike in 2008? \\
\bottomrule
\end{tabular}
\end{table*}

\end{appendices}

\end{document}